\documentclass[10pt,letterpaper]{article}
\usepackage{opex3}
\usepackage{cite}
\pdfoutput=1 %
\usepackage{amsmath}
\usepackage{amssymb}

\usepackage{graphicx}
\usepackage{caption}
\usepackage{subcaption}

\usepackage{algorithm}
\usepackage{algpseudocode}

\usepackage{tikz}
\usepackage{pgfplots}
\usetikzlibrary{plotmarks}
\usetikzlibrary{calc}
\usetikzlibrary{patterns}
\usetikzlibrary{external}

\newcommand{\filterBand}{$[0.5 \text{Hz},5\text{Hz}]$}

\newcommand{\algo}{distancePPG}
\newcommand{\mrc}{MRC}
\newcommand{\pulseox}{pulse oximeter}
\newcommand{\motion}{region based motion tracking}
\newcommand{\Motion}{Region based motion tracking}
\newcommand{\oldMethod}{face averaging method}

\newcommand{\Algo}{DistancePPG}

\newcommand{\DSNRStatic}{4.1}
\newcommand{\DSNRDark}{6-7}

\newcommand{\DoldHRMD}{-0.40}

\newcommand{\DoldHRMDplusSD}{3.7}
\newcommand{\DoldHRMDminusSD}{-4.5}

\newcommand{\DmrcHRMD}{-0.02}

\newcommand{\DmrcHRMDplusSD}{0.72}
\newcommand{\DmrcHRMDminusSD}{-0.75}

\newcommand{\DSNRMotion}{4.5}

\newcommand{\DoldHRMDmotion}{7.17}

\newcommand{\DoldHRMDplusSDmotion}{33.04}
\newcommand{\DoldHRMDminusSDmotion}{-18.70}

\newcommand{\DmrcHRMDmotion}{0.48}

\newcommand{\DmrcHRMDplusSDmotion}{6.70}
\newcommand{\DmrcHRMDminusSDmotion}{-5.73}

\newcommand{\DmrcHRMDreadwatch}{0.17}

\newcommand{\DmrcHRMDplusSDreadwatch}{4.86}
\newcommand{\DmrcHRMDminusSDreadwatch}{-4.52}

\newcommand{\DSNRGainLightW}{1.9}
\newcommand{\DSNRGainLightB}{6.5}

\newcommand{\DSNRICAStatic}{3.57}

\newcommand{\DSNRmrcStatic}{4.79}

\newcommand{\DMeanHRErrorICAStatic}{-0.17 \pm 1.21}

\newcommand{\DMeanHRErrormrcStatic}{-0.19 \pm 1.10}

\newcommand{\DRMSEIBIICAStatic}{58.14}

\newcommand{\DRMSEIBImrcStatic}{36.84}

\newcommand{\DSNRICAMotion}{-5.32}

\newcommand{\DSNRmrcMotion}{-4.62}

\newcommand{\DMeanHRErrorICAMotion}{8.09 \pm 22.44}

\newcommand{\DMeanHRErrormrcMotion}{14.52 \pm 12.82}

\begin{document}

\title{\Algo: Robust non-contact vital signs monitoring using a camera}

\author{Mayank Kumar$^{*}$, Ashok Veeraraghavan and Ashutosh Sabharval}

\address{Electrical and Computer Engineering, RICE University, Houston, TX, USA\\
}

\email{$^*$mk28@rice.edu} 



\begin{abstract}
Vital signs such as pulse rate and breathing rate are currently measured using contact probes. But, non-contact methods for measuring vital signs are desirable both in hospital settings (e.g. in NICU) and for ubiquitous in-situ health tracking (e.g. on mobile phone and computers with webcams). Recently, camera-based non-contact vital sign monitoring have been shown to be feasible. However, camera-based vital sign monitoring is challenging for people with darker skin tone, under low lighting conditions, and/or during movement of an individual in front of the camera.  In this paper, we propose \algo, a new camera-based vital sign estimation algorithm which addresses these challenges. \Algo\ proposes a new method of  combining skin-color change signals from different tracked regions of the face using a weighted average, where the weights depend on the blood perfusion and incident light intensity in the region, to improve the signal-to-noise ratio (SNR) of camera-based  estimate. One of our key contributions is a new automatic method for
determining the weights based only on the video recording of the subject. The gains in SNR of camera-based PPG estimated  using \algo\ translate into reduction of the error in vital sign estimation, and thus expand the scope of camera-based vital sign monitoring to potentially challenging scenarios. Further, a dataset will be released,  comprising of synchronized video recordings of face and \pulseox\ based ground truth recordings from the earlobe for people with different skin tones, under different lighting conditions and for various motion scenarios. 
\end{abstract}


\bibliographystyle{osajnl}

\section{Introduction}

Regular and noninvasive measurement of vital signs such as pulse rate (PR), breathing rate (BR), pulse rate variability (PRV), blood oxygen level (SpO2) and blood pressure (BP) are important both in-hospital and at-home due to their fundamental role in the diagnosis of health conditions and monitoring of well-being.  Currently, the gold standard techniques to measure the vital signs are based on contact sensors such as ECG probes, chest straps, pulse oximeters and blood pressure cuffs. However, contact-based sensors are not convenient in all scenarios, e.g. contact sensors are known to cause skin damage in pre-mature babies during their treatment in a neonatal intensive care unit (NICU).

Non-contact methods for vital sign monitoring using a camera has been recently shown  to be feasible \cite{verkruysse_remote_2008,poh_advancements_2011,sun_motion-compensated_2011}. Being non-contact, camera-based vital sign monitoring  have many applications --- from monitoring newborn babies in the NICU to in-situ continuous monitoring in everyday scenarios like working  in front of a computer. However, camera-based vital sign monitoring does not perform well for subjects having darker skin tones and/or under low lighting conditions as was highlighted in  \cite{aarts_non-contact_2013}.  Furthermore, current known algorithms require a person to be nearly at rest facing a camera to ensure reliable measurements. In this paper, we address the challenge of reliable vital sign estimation for people having darker skin tones, under low lighting conditions and under different natural motion scenarios to expand the scope of camera-based vital sign monitoring.  

Photoplethysmography (PPG) is an optical method to measure cardiac-synchronous  blood volume change in body extremities  such as the face, finger and earlobe. As the heart pumps blood, the volume of blood in the arteries and capillaries changes by a small amount in sync with the cardiac cycle. The change in blood volume in the arteries and capillaries underneath the skin leads to small change in the skin color.  The goal of a  camera-based vital sign monitoring system is to estimate the PPG waveform which is proportional to these small changes in skin color. Vital signs such as PR, PRV, SpO2 and BR can be derived from a well-acquired PPG waveform. 

The two major challenges in estimating PPG using a camera are: (i) extremely low signal strength of the color-change signal, particularly for darker skin tones and/or under low lighting conditions, and (ii)  motion artifacts due to an individual's movement in front of the camera.  Our main contribution in this paper is a new algorithm, labeled as \emph{\algo}, that improves the signal strength of camera-based PPG signal estimate, with following three key contributions.
\begin{itemize}
\item A new method to improve the SNR of camera-based PPG signal by combining the color-change signals obtained from different regions of the face  using a weighted average. 
\item A new automatic method for determining the weights based only on the video recording of the subject. The weights capture the effect of incident light intensity and blood perfusion underneath the skin on the strength of color-change signal obtained from a region. 
\item A method to track different regions of the face separately as the person moves in front of the camera using a combination of a deformable face tracker \cite{saragih_deformable_2011} and KLT (Kanade-Lucas-Tomasi) feature tracker \cite{lucas_iterative_1981, tomasi_detection_1991} to extract the PPG waveform under motion. 
\end{itemize}

For different skin tones (pale white to brown), the \algo~algorithm improves the signal to noise ratio (SNR)  of the estimated PPG signal on an average by $\DSNRStatic$~dB compared to prior methods \cite{verkruysse_remote_2008,poh_advancements_2011,sun_motion-compensated_2011}. Particularly, the improvement in SNR for non-white skin tone is $\DSNRDark$~dB. We have evaluated PPG estimation under three natural motion scenarios: (i) reading content on screen, (ii) watching video, and (iii) talking. \Algo\ improves the SNR of estimated camera PPG in these scenario by $\DSNRMotion$~dB on an average. Further, it improves the SNR of camera based PPG by as much as $\DSNRGainLightB$~dB under low lighting condition when compared to prior methods \cite{verkruysse_remote_2008,poh_advancements_2011,sun_motion-compensated_2011}. 

The improvement in SNR of camera-based PPG signal estimated using \algo\ reduces the error in pulse rate estimates in various scenarios. Using \algo, the  mean bias (average difference between ground truth \pulseox\ derived PR and camera-based PPG derived PR)   $\bar{d}$ is $\DmrcHRMD$ beats per minute (bpm) with $95\%$ limit of agreement (mean bias $\pm 1.96$ standard deviation of the difference) between $\DmrcHRMDminusSD$ to $\DmrcHRMDplusSD$ bpm for  $12$ subjects having skin tones ranging from fair white to brown/black. When using prior methods, the corresponding performance numbers are $\bar{d}$ = $\DoldHRMD$ with  $95\%$ limit of agreement between $\DoldHRMDminusSD$ to $\DoldHRMDplusSD$.  Under three motion scenario of reading, watching, and talking for $5$ subjects of varying skin tones, the mean deviation is $\bar{d}$ = $\DmrcHRMDmotion$ bpm with $95\%$ limit of agreement between $\DmrcHRMDminusSDmotion$ to $\DmrcHRMDplusSDmotion$ bpm using \algo. Using prior methods, the corresponding performance numbers are $\bar{d} = \DoldHRMDmotion$ bpm with $95\%$ limit of agreement between $\DoldHRMDminusSDmotion$ to $\DoldHRMDplusSDmotion$ bpm.  Further, \algo\ reduces the root mean square error (RMSE) in pulse rate variability estimation below $16$~ms for non-black skin tones using a $30$~fps camera.

\subsection{Prior work}
\label{sec:priorArt}

Over the past decade, several researchers have worked on measuring vital signs such as PR, PRV, BR, and SpO2 using a camera  \cite{wieringa_contactless_2005,humphreys_noncontact_2007,verkruysse_remote_2008,poh_non-contact_2010,poh_advancements_2011,sun_motion-compensated_2011,holton_signal_2013}. Initially \cite{wieringa_contactless_2005,humphreys_noncontact_2007},
external arrays of LEDs at $760$nm and $880$nm were used to illuminate a region of tissue for measuring PR  using a monochrome CMOS camera. It was shown later \cite{verkruysse_remote_2008} that PR and BR can also  be determined using simply a color camera and ambient illumination. The authors in \cite{verkruysse_remote_2008} found that the face is the best region to extract PPG signal because of better signal strength. They also reported  that the \emph{green channel} of the RGB camera perform better than red and blue channel for detecting PR and BR. The fact that the green channel perform better is expected as the absorption spectra of hemoglobin (Hb) and oxyhemoglobin (HbO$_2$), two main constituent chromophores in blood, peaks in the region around $520-580$~nm, which is essentially the passband range of the green filters in color cameras. 

Further in \cite{poh_non-contact_2010, poh_advancements_2011}, the authors used a color webcam under ambient illumination to detect simultaneous PR, PRV and BR of multiple people in a video by using automatic face detection to define the face region. They used blind source separation (BSS) to decompose the three camera channels (red, green, blue) into three independent source signals using independent component analysis (ICA) algorithm, and extracted the PPG signal from one of the independent sources. More recently, authors in \cite{mcduff_improvements_2014} demonstrated that cyan, orange and green (COG) channels work better than red, green, and blue (RGB) channels of a camera for camera-based vital sign estimation. One possible explanation for better performance of cyan-orange-green (COG) channels could be the higher overlap between the passband of cyan ($470-570$~nm), green ($520-620$~nm), and orange ($530-630$~nm) with the peak in the absorption spectra of Hb and HbO$_2$ ($520-580$~nm).

Most of the past work, however, did not report how camera-based PPG performs  on individuals with different skin tones, under low lighting conditions, and for different motion scenarios. It is well-known that the higher amount of \emph{melanin} present in darker skin tones absorbs a significant amount of incident light, and thus degrades the quality of the camera-based PPG signal, making the system ineffective for extracting vital signs in darker skin tones. Recently, a pilot study conducted in NICU for monitoring pulse rate using a camera-based method  showed difficulty under low lighting conditions and/or under infant motion \cite{aarts_non-contact_2013}. 

To counter the motion challenge, the authors in~\cite{poh_non-contact_2010} used automatic face detection in consecutive frames to track the face. But, they reported difficulties in continuously detecting faces under motion  due to large false-negatives.  Another method is to compute 2D shifts in face location between consecutive frames \cite{sun_motion-compensated_2011} using image correlation to model the motion. By simply computing the global 2D shifts, one can only capture the basic translation motion of face, and it is difficult to compensate for more natural motion like turning or tilting of the face, smiling or talking, generally found in in-situ scenarios.

\section{Background and problem definition}
For a camera-based PPG estimation we record the video of a person facing a camera, and the objective is to develop an algorithm to estimate the underlying PPG signal $p(t)$ using the recorded video.  The recorded video is in the form of intensity signal $V(x,y,t)$ comprising of sequences of frames $V(x,y,t=1,2,3 \cdots)$. Each frame of the video records the intensity level of the light reflected back from the face over a two dimensional grid of pixel $(x,y)$ in the camera sensor. If the camera sensor has multiple color channels (e.g. red, green, blue), one can get separate intensity signals corresponding to each channel (e.g. $V_r(x,y,t),V_g(x,y,t),V_b(x,y,t)$). 

In general, the measured intensity of any reflected light can be decomposed into two components: (i) intensity of illumination, and (ii) reflectance of the surface (skin), i.e. 
\begin{equation}
V(x,y,t) = I(x,y,t)R(x,y,t). 
\end{equation}
The illumination intensity corresponds to the intensity of ambient or any dedicated light falling on the face. For PPG estimation, it is generally assumed that the light intensity remains same over the PPG estimation window (typically $5-60$~sec in past works). The skin reflectance $R(x,y,t)$ is equal to the fraction of light reflected back from the skin and consists of two level of light reflectance: (i)~surface reflection, and (ii)~subsurface reflection or backscattering.  

A large part of the light incident on face gets reflected back from the surface of the skin, and is characterized by the skin's bidirectional reflectance distribution function (BRDF). Remaining part of the incident light goes underneath the skin surface and is absorbed by the tissue and the chromophores (Hb, HbO$_2$) present in blood inside arteries and capillaries. The volume of blood in the arteries and capillaries changes with each cardiac cycle and thus the level of absorption of light changes as well. Since PPG signal, by definition, is proportional to this cardio-synchronous pulsatile blood volume change in the tissue  \cite{allen_photoplethysmography_2007}, one can estimate PPG signal $p(t)$ by estimating these small changes in subsurface light absorption. Thus, the camera-based PPG  signal is estimated by extracting small variations in the subsurface component of  skin reflectance $R(x,y,t)$. 

\subsection{A prototypical solution}
Since the incident light intensity $I(x,y)$ is assumed to be constant over PPG estimation time window, any temporal change in the intensity of the light reflected back from the face region will be proportional to the changes in the reflectance of the skin surface $R(x,y,t)$. Generally, these temporal changes in recorded intensity will be dominated by changes in surface reflection component unrelated to the underlying PPG signal of interest. As a first step for camera-based PPG estimation, one can  spatially average the recorded intensity level $V(x,y,t)$ over all the pixels in the face region to yield a measurement point per frame in the video \cite{verkruysse_remote_2008,poh_non-contact_2010}. The basic idea is that by averaging the intensity signal, the incoherent changes in surface reflection component over all the pixels inside the face will cancel out, and the coherent changes in subsurface reflection component due to blood volume changes will add up to give an estimate  $y_{\text{face}}(t)$. The spatially averaged intensity signal $y_{\text{face}}(t)$ would be proportional to the  changes in the subsurface reflectance component, and thus to the underlying PPG signal of interest. One generally filters $y_{\text{face}}(t)$ signal between $0.5$~Hz to $5$ Hz (frequency band of interest) to extract the PPG signal.

\section{Key challenges and insights for camera-based PPG estimation}

\begin{figure}[t]
\centering
	\begin{subfigure}[b]{0.35\textwidth}
	
	\begin{tikzpicture}
		\node[anchor=north] (img) at (0,0)
		{\includegraphics[width=1.0\textwidth]{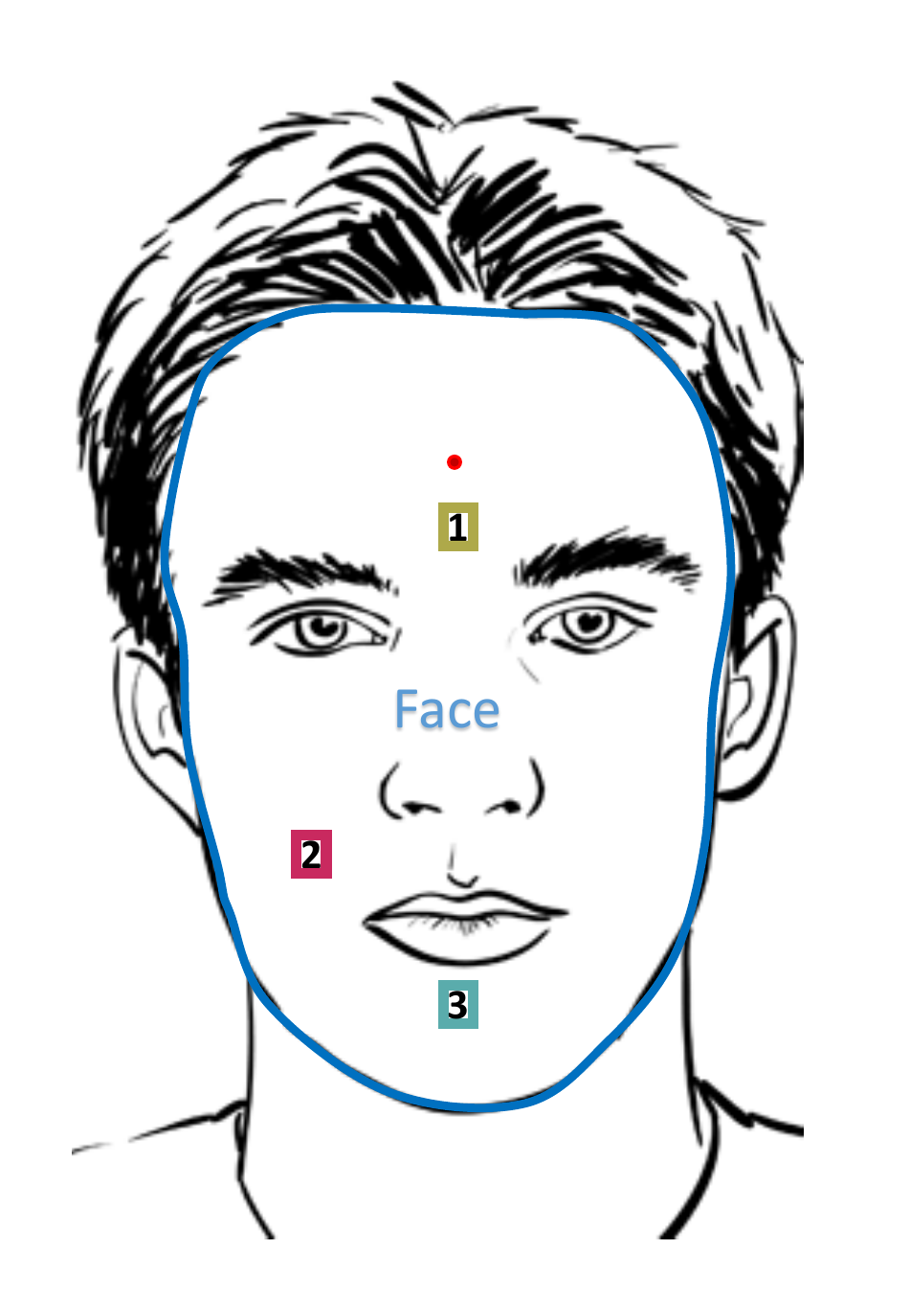}};
		\node[text width = 2.1in, align=center, below, anchor=north] (text) at ($(img.south)+(0,+0.05)$)
		{PPG signal strength varies over different 
		regions in the face and depends on intensity of incident light $I(x,y)$ in the region and on the blood perfusion
		underneath the skin.};
		
	\end{tikzpicture}
	
	\label{fig:faceROI}
		
	\end{subfigure}~\begin{subfigure}[b]{0.65\textwidth}
		\centering
		\begin{tikzpicture}
			\begin{axis}[
					name=raw,
					width=3in,
					height=1in,
					xmin=-1,
					xmax=20,
					xtick={\empty},
					ymin=-10,
					ymax=10,
					ytick={-5,5},
					axis x line = none,
					axis y line = left,
					y axis line style=-,
					]
					\addplot [color=red,solid, thin] table[x=time,y=raw]{result/RegionPPG.txt};
				\end{axis} 
				\node[anchor=west] (raw_text) at ($(raw.east)+(0.1,0.0)$)
				{$y_{\text{raw}}(t)$};
				\begin{axis} [
					name=face,
					at={($(raw.south)-(0,0.2cm)$)},
					width=3in,
					height=1in,
					anchor=north,
					xmin=-1,
					xmax=20,
					xtick={\empty},
					ymin=-.5,
					ymax=.5,
					ytick={-0.25,0.25},
					axis x line = none,
					axis y line = left,
					y axis line style=-,	
					]
					\addplot [color=cyan,solid, thin] table[x=time,y=face]{result/RegionPPG.txt};
				\end{axis} 
				\node[anchor=west] (face_text) at ($(face.east)+(0.1,0.0)$)
								{$y_{\text{face}}(t)$};		
				\begin{axis} [
					name=RA,
					at={($(face.south)-(0,0.2cm)$)},
					width=3in,
					height=1in,
					anchor=north,
					xmin=-1,
					xmax=20,
					xtick={\empty},
					ymin=-2,
					ymax=2,
					ytick={-1,1},
					ylabel=Camera Intensity Scale,
					axis x line = none,
					axis y line = left,
					y axis line style=-,	
					]
					\addplot [color=olive,solid, thin] table[x=time,y=R1]{result/RegionPPG.txt};
				\end{axis} 
					\node[anchor=west] (RA_text) at ($(RA.east)+(0.1,0.0)$)
												{$y_{1}(t)$};
				\begin{axis} [
					name=RB,
					at={($(RA.south)-(0,0.2cm)$)},
					width=3in,
					height=1in,
					anchor=north,
					xmin=-1,
					xmax=20,
					xtick={\empty},
					ymin=-2,
					ymax=2,
					ytick={-1,1},
					axis x line = none,
					axis y line = left,
					y axis line style=-,	
					]
					\addplot [color=purple,solid, thin] table[x=time,y=R2]{result/RegionPPG.txt};
				\end{axis} 
				\node[anchor=west] (RB_text) at ($(RB.east)+(0.1,0.0)$)
												{$y_{2}(t)$};
				
				\begin{axis} [
					name=RC,
					at={($(RB.south)-(0,0.2cm)$)},
					width=3in,
					height=1in,
					anchor=north,
					xmin=-1,
					xmax=20,
					xtick={\empty},
					ymin=-2,
					ymax=2,
					ytick={-1,1},
					axis x line = none,
					axis y line = left,
					y axis line style=-,	
					]
					\addplot [color=teal,solid, thin] table[x=time,y=R3]{result/RegionPPG.txt};
				\end{axis} 
				\node[anchor=west] (RC_text) at ($(RC.east)+(0.1,0.0)$)
																{$y_{3}(t)$};
				\begin{axis} [
					name=ground,
					at={($(RC.south)-(0,0.25cm)$)},
					width=3in,
					height=1in,
					anchor=north,
					xmin=-1,
					xmax=20,
					axis x line = bottom,
					axis y line = left,
					y axis line style=-,
					ymin=-0.5,
					ymax=0.5,
					ytick = {\empty},
					xlabel=Time(s),]
					\addplot [color=gray,solid, thin] table[x=time,y=ground]{result/RegionPPG.txt};
				\end{axis}
				\node[anchor=west] (RA_text) at ($(ground.east)+(0.1,0.0)$)
									{$z_{\text{pulseOx}}(t)$};

		\end{tikzpicture}
	
	\end{subfigure}
\caption{Camera-based PPG estimation: strength of PPG signal obtained from different skin patch in the face is different due to spatial variations in blood perfusion. \Algo\ combines the average pixel intensity signal $y_i(t)$ from different regions of the face using a weighted average to maximize the SNR of the overall estimate of the PPG signal, $y_{\text{raw}}(t)$ shows the PPG signal obtained from a single pixel in the forehead (marked in red) for comparison. }
\label{fig:ROI}
\end{figure}

\textbf{Challenge 1: Very low signal strength}: PPG signal extracted from camera video  have low signal strength. This is because the skin vascular bed contains a very small amount of blood (only $2-5\%$ of total blood in the body), and the blood volume itself experiences only a small ($5\%$) change in sync with the cardiovascular pulse  \cite{hu_opto-physiological_2013}. Thus, the change in subsurface skin reflectance due to cardio-synchronous changes in blood volume is very small. The small change in subsurface reflectance results in very small change in light intensity recorded using a camera placed at a distance. On the other hand, the change in surface reflection component due to small movement of person is very large. For example, see Figure~\ref{fig:ROI} where the top plot (red) shows recorded intensity changes in a single pixel marked on the forehead ($y_{\text{raw}}(t)$). When compared to the ground truth \pulseox\ signal (bottom, $z_{\text{pulseox}}(t)$), it is clear that the raw intensity change variations in $y_{\text{raw}}(t)$ are unrelated to the underlying PPG signal and is dominated by a significant amount of surface reflectance changes

To estimate small changes in subsurface reflectance, a general idea used in most past work is to spatially average the recorded intensity level over the face region. For example, see the plot of $y_{\text{face}}(t)$ in Figure~\ref{fig:ROI}. Clearly,  $y_{\text{face}}(t)$ is a better estimate of the PPG signal as is evident by comparing it with the ground truth \pulseox\ signal (bottom). The amplitude of $y_{\text{face}}(t)$ is within $\pm1$ in camera intensity scale (all these signals are filtered between \filterBand\ and so they are centered around $0$). This is at the level of the quantization step of $8-10$ bit ADC present in cameras and thus camera-based PPG estimate is corrupted with large quantization noise.

Now, let us suppose we spatially average only the pixels within the square blocks $1,2,3$ marked on the face ($20\times20$~px in size) in  Figure~\ref{fig:ROI}. By comparing the plot of $y_{1}(t)$, $y_{2}(t)$, and $y_3(t)$, we can clearly see that there are differences in the quality of estimated color change (or PPG) signal from these regions ($y_1(t)$ is a better estimate than $y_2(t)$ which is better than $y_3(t)$). Past work has not fully exploited the fact that the quality of PPG estimates obtained from different regions of the face varies significantly across the face.

Strength of the PPG signal extracted from a patch of the imaged skin surface would depend on the intensity of light $I(x,y)$ incident on that patch and on the total amount of blood volume change underneath the skin patch. The total amount of blood volume change will depend on the blood perfusion in the region, which in turn will be determined by the density of blood carrying capillaries in that region.  When pixel intensity is averaged over the whole face region to get $y_{\text{face}}(t)$, one ends up also including skin patches that have very limited blood perfusion, and hence contribute more noise than signal to the  overall estimate of the camera-based PPG signal. 

Thus, there is a need for an automated method to find out which regions in the face can be used to estimate better PPG signal and which regions should be rejected because they contribute more noise than signal to the overall PPG estimate. This is a challenging task because we do not have access to the ground truth PPG signal, and thus cannot determine the quality of PPG estimation from different regions in the face.  Further, it will also be desirable to combine the PPG signal obtained from different regions of the face in a manner that the overall estimate has maximum signal to noise ratio (SNR). 

\textbf{Challenge 2: Motion artifact}:
Until this point we have assumed that the person facing the camera is static (we will assume that the camera is static throughout this paper). If the person moves in front of camera, e.g. tilting, smiling, shifting, talking etc, one needs to at the very least track the whole face to extract the PPG signal.  As the face is not a rigid body, in a sense that different regions within the face can move separately (e.g. cheek muscles during talking, smiling etc), it would be ideal to track different non-rigid regions of the face independently.

Even if we could track different regions within the face faithfully, there are other challenges in faithful PPG estimation under motion. First, most light sources have a spatial illumination variations $I(x,y)$, and when a tracked region moves in space, it leads to change in incident light intensity in that region over time. Such changes violate the assumption that incident light intensity is constant over PPG estimation window and corrupt the  PPG estimate even when the tracking works perfectly. 

Second, even very small motion (e.g. slight rotation) of person's face relative to camera can lead to change in incident light direction (from light source to skin surface) and  reflected light direction (from skin surface to camera sensor). Such small changes in light direction can lead to large changes in skin surface reflectance which is characterized by the highly non-linear BRDF of the skin surface. This large change in surface reflectance can completely overwhelm the measurement of small changes in subsurface reflectance due to changes in blood volume in sync with the cardiac cycle (our PPG signal). 

The fact that blood volume change underneath the skin causes very small changes in the intensity of reflected light signal $V(x,y,t)$ recorded by the camera can be used to our advantage. For any region (skin patch) inside the face to be useful in estimating the PPG waveform, the corresponding averaged pixel intensity signal from that region should not change by very large magnitude within the PPG estimation window. Any large change in intensity is mostly due to changes in incident light intensity $I(x,y)$ or surface reflectance,  and would be much larger than the intensity changes we are interested in. So, it would be wise to identify such bad regions, and reject them completely till the time they are contributing large artifacts.

\section{\Algo: camera based PPG estimation algorithm}
\begin{figure}
\centering

\begin{tikzpicture}
\node[inner sep=0pt] (step1) at (0,0)
    {\includegraphics[height=1.2in,width=.25\textwidth]{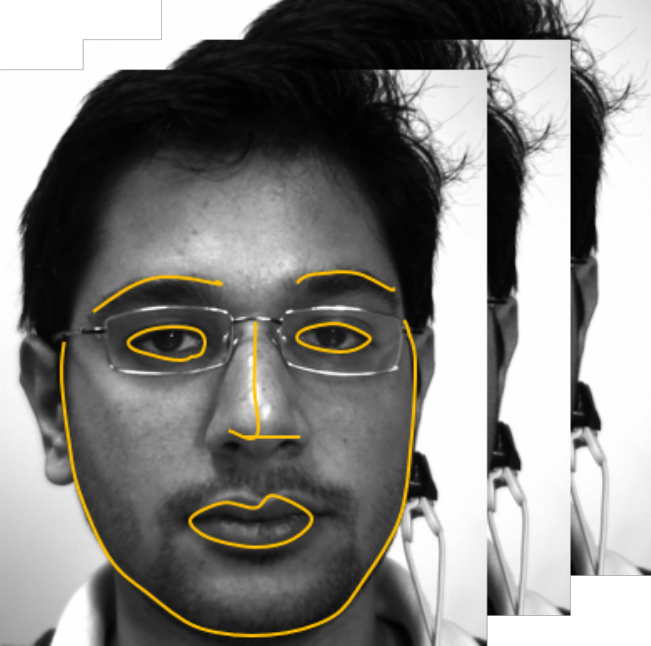}};
\node[text width = 1.4in, align=center, below] at ($(step1.south)+(-0.25,-0.5)$) {\textbf{Step 1:} Input: Green channel video of a person, landmark points around eyes, nose, mouth in face detected};

\node[inner sep=0pt] (step2) at (3.3,0)
    {\includegraphics[height=1.2in,width=.2\textwidth]{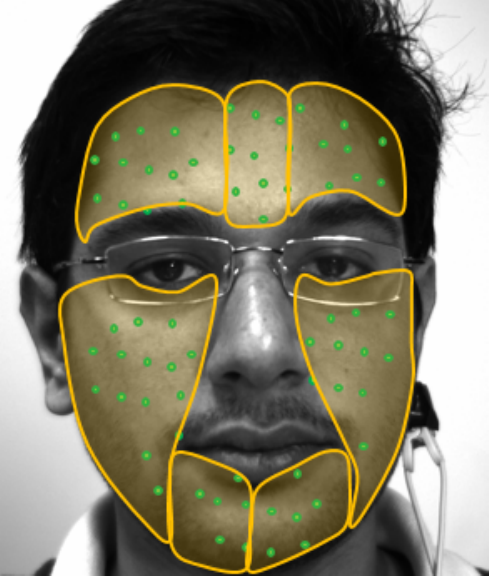}};

\node[text width = 1.4in, align=center, below] at ($(step2.south)+(-0.1,-0.5)$) {\textbf{Step 2:} Face is divided into seven regions, each region tracked using KLT, motion modeled using rigid affine fit};

\node[inner sep=0pt] (step3) at (6.51,0)
     {\includegraphics[height=1.2in,width=.2\textwidth]{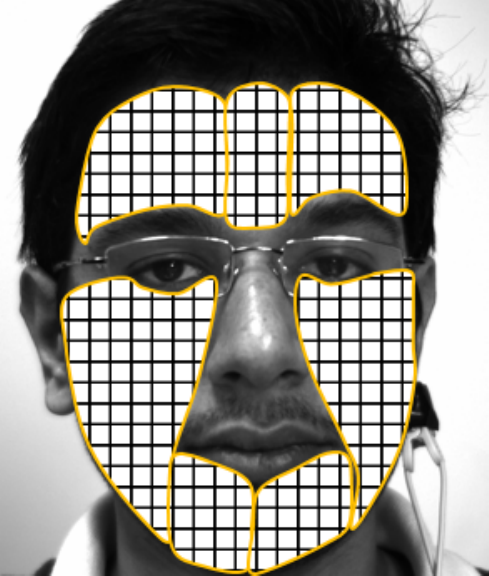}};

\node[text width = 1.4in, align=center, below] at ($(step3.south)+(0.1,-0.5)$) {\textbf{Step 3:} Each tracked region is divided into $20$x$20$ pixel block, avg. pixel intensity $y_i(t)$ computed from each ROI };

\node[inner sep=0pt] (step4) at (9.75,0)
    {\includegraphics[height=1.2in,width=.18\textwidth]{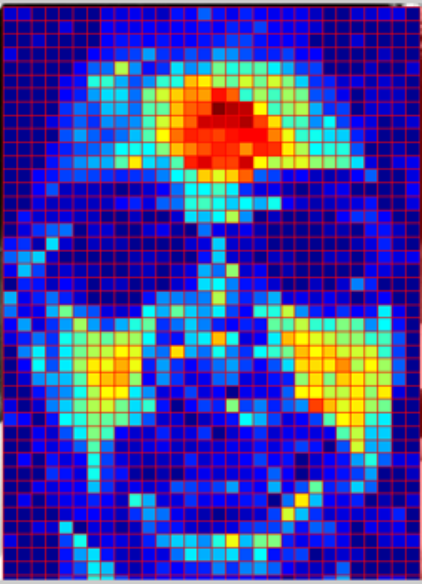}};

\node[text width = 1.3in, align=center, below] at ($(step4.south)+(0.3,-0.5)$) {\textbf{Step 4:} Goodness metric $G_i$ computed for each ROI, overall camera PPG estimated using weighted average };

\draw[->,thick] (step1.east) -- (step2.west);
\draw[->,thick] (step2.east) -- (step3.west);
\draw[->,thick] (step3.east) -- (step4.west);
\draw [-,dashdotted] ($(step1.north west)+(-0.25,0.25)$) -- ($(step1.south west)+(-0.25,-0.25)$) -- ($(step2.south east)+(+0.2,-0.25)$) -- ($(step2.north east)+(+0.2,+0.25)$) -- ($(step1.north west)+(-0.25,0.25)$);

\draw [-,dashdotted] ($(step3.north west)+(-0.2,0.25)$) -- ($(step3.south west)+(-0.2,-0.25)$) -- ($(step4.south east)+(+0.25,-0.25)$) -- ($(step4.north east)+(+0.25,+0.25)$) -- ($(step3.north west)+(-0.2,0.25)$);

\node[text width = 2in, align=center, above] at ($(step1.north)+(1.8,0.5)$) {\textbf{\Motion}};

\node[text width = 2in, align=center, above] at ($(step3.north)+(1.8,0.5)$) {\textbf{\mrc\ algorithm}};

\end{tikzpicture}

\caption{Overall steps involved in \algo\ algorithm for estimating camera-based PPG.}
\label{fig:blockDiag}
\end{figure}

We first propose a camera-based PPG signal acquisition model which captures the effect of incident light intensity, surface and subsurface skin reflectance, and camera noise on PPG signal estimation. Based on the proposed signal acquisition model, we develop  \algo\ estimation algorithm that comprises of two subparts: i)~\emph{\mrc}\ (maximum ratio combining) algorithm to combine the average pixel intensity signal extracted from different regions of the face to improve the overall signal to noise ratio (SNR) of the final PPG estimate, and (ii)~\emph{\motion}\ algorithm that keeps track of different regions of the face as the person moves in front of the camera. Overall steps  involved in \algo\ algorithm are shown in Figure~\ref{fig:blockDiag}.

\subsection{Camera-based PPG signal acquisition model} \label{sec:PPG_acquisition_model}
Given the intensity signal $V(x,y,t)$ of a person facing a camera, we assign the set of pixels imaging the face into many small regions of interest (ROIs) denoted by the set $\mathcal{R}$. These ROIs are small enough that the blood perfusion within them can be assumed to be constant. Let $y_i(t)$ be the spatial average of the intensity of the pixel within the ROI $\mathcal{R}_i$ at time $t$, $i\in\{1,2\cdots,n\}$ is used to index the elements of the set $\mathcal{R}$. Also, the spatial illumination variation $I(x,y)$ inside an ROI $\mathcal{R}_i$ can be assumed to be constant, and will be denoted as $I_i$. Then, $y_i(t)$ can be modeled as

\begin{equation}
y_i(t) = \overbrace{I_i}^{\text{Illumination}}\overbrace{\left(\underbrace{\alpha_i\cdot p(t)}_{\text{subsurface reflectance}} +\underbrace{b_i}_{\text{Surface reflectance}}\right)}^{\text{Reflectance}}+ \overbrace{q_i(t)}^{\text{Quantization noise}},
\end{equation}
where $I_i$ is the incident light intensity in ROI $\mathcal{R}_i$, $\alpha_i$ is the strength of blood perfusion, $b_i$ is the surface reflectance from the skin in region  $\mathcal{R}_i$, and $q_i(t)$ is the camera quantization noise. When incident illumination $I_i$ falls on skin ROI  $\mathcal{R}_i$, a large fraction of it ($b_i$) is reflected back by the skin surface and does not contain any pulsatile component related to blood volume change. Some part of the incident light penetrates beneath the skin surface, and gets modulated by the pulsatile blood volume change waveform $p(t)$, due to light absorption, before reflecting back (back-scattering).  Here, $\alpha_i$ represent the strength of modulation of light backscattered from the subsurface due to the pulsatile blood volume change. The parameter $\alpha_i$ will  primarily depends on the average blood perfusion in the selected ROI, and thus varies over different regions of the face and is also different for different individuals. 

The parameter $\alpha_i$ will also depend on the wavelength of incident light since the absorption spectra of Hb and HbO$_2$, two major chromophores present in blood, depends on wavelength of light. In this work, we only consider camera's green channel that coincides with the peak in the absorption spectra of Hb and HbO$_2$, This choice  provides best performance in PPG estimation as was also highlighted by other researchers \cite{verkruysse_remote_2008}. So, we did not explicitly incorporate wavelength dependence of $\alpha_i$ in the model equation.

Our goal is to extract $p(t)$ from the measurements $y_i(t)$ in the presence of large surface reflectance  $b_i$ which constitutes a dominant portion of the signal captured by the camera sensor. Quantization noise is also significant as our signal of interest $p(t)$  varies by a small amount relative to the surface reflection. If the person moves in front of the camera, then all the ROIs in set  $\mathcal{R}$ need to be tracked. 

\subsection{\mrc\ algorithm} \label{sec:MRC}
Given the intensity signal $V(x,y,t)$ of a person facing a camera, we first compute all the $y_i(t)$ corresponding to the ROI $\mathcal{R}_i$ inside the face. Each $y_i(t)$ contain different strength of the underlying PPG signal $p(t)$, and different surface reflection component $I_i\cdot b_i$. Then, we temporally filter all $y_i(t)$ using a bandpass filter~\filterBand~  to reject the out of band component of skin surface reflection ($I_i\cdot b_i$) and other noise outside the band of interest to obtain $\hat{y}_i(t)$.

Based on the camera-based PPG signal acquisition model in Section~\ref{sec:PPG_acquisition_model}, the filtered signals from different regions of the face can be written as 
\begin{eqnarray} \label{ROI_PPG}
\hat{y}_1(t) & = & A_1 p(t) + w_1(t), \nonumber \\
\hat{y}_2(t) & = & A_2 p(t) + w_2(t), \nonumber \\
\vdots \nonumber \\
\hat{y}_n(t) & = & A_n p(t) + w_n(t),
\end{eqnarray} 
where $i\in \{1,2, \cdots, n\}$ represents the corresponding ROI number in $\mathcal{R}$.  Here, $A_i$ denote the strength of the underlying PPG signal in region $\mathcal{R}_i$ and is determined both by the strength of modulation $\alpha_i$ and the incident light illumination $I_i$.  Further, $w_i(t)$ denote the noise component due to the camera quantization, unfiltered surface reflection and motion artifacts. 

Here, $\hat{y}_1(t), \hat{y}_2(t), \cdots, \hat{y}_n(t)$ can be considered as different channels that receive different strength of the  same desired signal $p(t)$, and have different level of noise. We can combine all these different channels using a weighted average
\begin{equation}
\hat{p}(t) = \sum_{i=1}^nG_i\hat{y}_i(t). 
\end{equation}
The weights for each channel can be determined based on the the idea of maximum ratio diversity~\cite{brennan_linear_2003}. The maximum ratio diversity algorithm states  that the assigned weights should be proportional to the root-mean-squared (RMS) value of the signal component, and inversely proportional to the mean-squared noise in that channel, in order to maximize the signal-to-noise ratio of the overall camera-based PPG estimate $\hat{p}(t)$; mathematically, 
\begin{equation}
G_i = \frac{A_i}{\big\|w_i(t)\big\|^2}.
\end{equation}
As both $A_i$ and $w_i$ are unknown in our case,  we need to develop an alternative method to estimate these weights $G_i$. For ease of reference, we label $G_i$ as \emph{goodness metric} for region $\mathcal{R}_i$ from hereon. The maximum ratio diversity algorithm assumes that the signal component $A_ip(t)$ is locally coherent among all channel, i.e. there is no time delay between PPG signals extracted from different ROIs, and the noise component $w_i(t)$ is uncorrelated. 

Generally speaking, PPG signal obtained from different regions of the skin would exhibit varying delays as the blood reaches these regions at different times. For example, there is a time lag of $80$ ms in PPG recorded between finger and toe \cite{nitzan_difference_2002}. However, the time lag between the PPG signal obtained from close-by regions are small, e.g. all the regions inside the face. Our own measurements show that the delay is less than $10$ ms between farthest point in face. As this delay falls within one sample period of normal cameras having frame rate of $30 - 60$~Hz, we can neglect such small delays for all practical purposes, and thus the signal component $A_ip(t)$ can be considered as locally coherent. 

As the amplitude of the camera-based PPG signal of interest $A_ip(t)$ is generally very small (within 1-2 bits of the camera ADC), we also reject regions which have unusually large signals. Large variations are mostly due to illumination change or motion artifacts. Thus, we reject all regions having amplitude greater than a threshold $A_{th}$. Our final estimate of the PPG signal over a time window of $T$ sec is given by
\begin{equation}\label{eg:PPG_estimate}
\hat{p}(t) = \sum_{i=1}^nG_i\hat{y}_i(t)I(\hat{y}_{\max,i}-\hat{y}_{\min,i}<A_{th}),
\end{equation}
where $I(\cdot)$ is the indicator function, $\hat{y}_{\max,i} = \max_{t \in [0,T]} \hat{y}_i(t)$, $\hat{y}_{\min,i} = \min_{t \in [0,T]} \hat{y}_i(t)$ is the maximum and minimum amplitude of the $\hat{y}_i(t)$ over a $T$ sec duration.

\subsection{Estimating goodness metric}
The PPG signal $p(t)$ has a fundamental frequency of oscillation equal to the pulse rate. Thus, the spectral power of the PPG signal is concentrated in a small frequency band around the PR. Moreover, the spectral power of the noise $w_i(t)$ present in $\hat{y}_i(t)$ will be distributed over the passband of the filter \filterBand.  Based on this spectral structure of the signal, we can estimate the goodness metric as a ratio of the power of recorded $\hat{y}_i(t)$ around the pulse rate (PR) to the power of the noise in the passband of the filter. 

Let $\hat{Y}(f)$ be the \emph{power spectral density} (PSD) of  $\hat{y}(t)$ over time duration $[0,T]$. Then goodness metric $G_i$ can be defined  as    
\begin{equation} \label{Goodness definition}
G_i(PR) = \frac{\int_{PR -b}^{PR+b}\hat{Y}_i(f)\mathrm{d}f}   {\int_{B_1}^{B_2}\hat{Y}_i(f)\mathrm{d}f-\int_{PR-b}^{PR+b}\hat{Y}_i(f)\mathrm{d}f},
\end{equation} 
where $[\text{PR}-b,\text{PR}+b]$ denote a small region around the pulse rate (PR) of the person, and $[B_1,B_2]$ is the passband of the bandpass filter (\filterBand). 

The definition of $G_i$ depends on the pulse rate which is still unknown. A two-step process can be followed to get a coarse estimate of pulse rate. First, we assume that all the $G_i$ are equal to $1.0$, and thus we compute a temporary coarse estimate of the PPG signal using Equation~\eqref{eg:PPG_estimate}. Second, we compute the frequency corresponding to the peak in the spectrum of this coarse estimate of PPG signal to get a coarse estimate of the pulse rate to be used in goodness metric definition. 
\begin{figure}[t]
\centering
\includegraphics[width=0.9\linewidth]{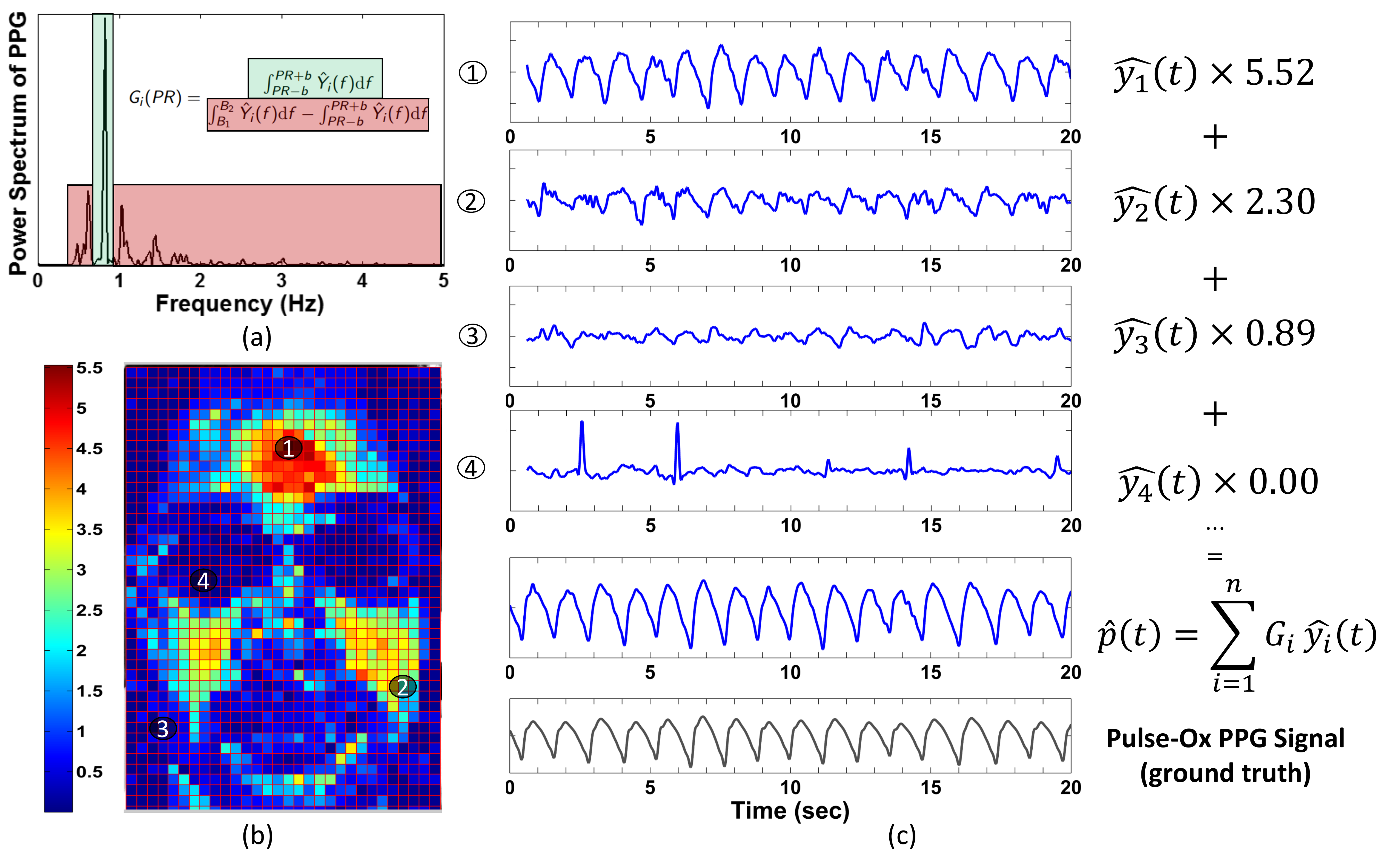}
\caption{Working of \mrc: (a) illustrates the goodness metric definition based on the area under the power spectrum density (PSD) of PPG, (b) shows a face with the goodness metric overlay, red regions have higher goodness metric, blue regions have lower goodness metric,  (c) shows the PPG signal extracted from four different regions marked on the face, also shows the weighted average camera PPG estimate  which is very similar in shape to pulse-ox PPG signal (ground truth) }
\label{fig:MRC_Algorithm}
\end{figure}

Figure~\ref{fig:MRC_Algorithm}(a) shows the PSD of acquired PPG signal and highlights how the goodness metric is defined based on the area under the PSD. Figure~\ref{fig:MRC_Algorithm}(b) shows a face with the goodness metric overlay. The goodness metric is based only on the recorded video of a person's face, and thus adapts to changes in blood perfusion for different people or changes in lighting conditions.  Regions shown as more red have higher goodness value, those shown in blue have lower goodness metric. Figure~\ref{fig:MRC_Algorithm}(c) shows the PPG signal extracted from four different regions marked on the face. When we compare these PPG signals with the \pulseox\ based ground truth, it is evident that the goodness metric correctly predicts the strength of the PPG signal extracted from these regions relative to the noise power. Thus, forehead (ROI 1, Goodness Metric = $5.52$) and cheek (ROI 2, Goodness Metric = $2.30$) regions give higher strength PPG signal estimates, whereas the region around mouth (ROI 3, Goodness = $0.89$) and eyes (ROI 4, Goodness = $0.00$) do not yield good signal. Since the ROI $4$ around eyes gives spiky signal due to eye movements, the signal amplitude in this region crossed the amplitude threshold $A_{th}$. Thus, the weight given to ROI $4$ is $0.00$, or equivalently that region is rejected by the algorithm.

\subsection{\Motion\ algorithm} \label{sec:motion}

In camera-based acquisition of PPG, even a slight motion of person in front of camera  can change the relative position of selected camera's ROI $\mathcal{R}_i$ and imaged skin portion. As our \mrc\ algorithm requires PPG extraction from all the different ROIs independently, we need to keep track of each selected ROI on the skin surface as the person moves in front of the camera. 

As a first step in tracking, we extract landmark location from the face like location of eyes, nose, mouth and the outer boundary  using a deformable face fitting algorithm described in \cite{saragih_deformable_2011}. Relative to these landmark locations, we then define seven planar regions in the face denoted by set $\mathcal{P}$ for tracking --- three on forehead, left and right cheek, and two on the chin. Particularly, we do not include regions around the mouth and eyes, as they exhibit non-rigid motion and are difficult to track and thus causes large motion artifacts. 

As a next step, to track these planar regions on the face across the video frames, we identify $M \simeq 50$ feature points inside each planar region. These feature points are selected so that they could be tracked well, and are known as \emph{good features to track} \cite{shi_good_1994} in computer vision literature.  We then use the Kanade Lucas Tomasi (KLT) feature tracker \cite{lucas_iterative_1981, tomasi_detection_1991} to track these features across the video. Afterwards, we compute a rigid affine fit for each planar region using the tracked feature points inside corresponding planar region to model its motion  between consecutive frames

We use more feature points than is minimally required for a rigid affine fit ($3$ feature points) as feature tracking is generally error prone due to feature points changing appearance or disappearing from camera view due to occlusion. We use the method proposed in \cite{kalal_forward-backward_2010} to automatically detect tracking failures based on the forward-backward error and do not consider erroneous points for affine model estimation. From the remaining well-tracked feature points, we use the random sample consensus (RANSAC) \cite{fischler_random_1981}  algorithm to compute a robust estimate of affine fit by considering only the inlier points for the model, and rejecting the outliers. 

We then divide these seven planar regions $\mathcal{P}$ on the face into many small PPG region of interest required by the \mrc\ algorithm. Each PPG ROI $\mathcal{R}_i$ is tracked across the video frames using the estimated affine motion model of the encompassing planar region. We  then combine the PPG signal from all the tracked ROI using the \mrc\ algorithm. After every $T$ seconds, we reset the KLT tracker and reinitialize the deformable face model.  All the steps for \motion\ algorithm  discussed above are summarized in Algorithm~\ref{motion}.

\begin{algorithm}
\caption{\algo~Region Tracker algorithm}\label{motion}
\begin{algorithmic}[]
\Procedure{TrackPPGRegion}{$V(x,y,t)$}\Comment{Track ROIs given a sequence of frames}
\State  $ t \gets 0$ \Comment{Current Frame Counter}
\While{$t \leq T_{total}$ } \Comment{$T_{total}$ is total number of frames}
\If{ $t \bmod T\times FPS = 0$ } \Comment{Restart after every $T$ sec. $FPS$ is Camera frame rate}
	\State LP $\gets$ \textsc{DeformableFace}($V(x,y,t)$) \Comment{LP are landmark points.}
	\State  $\{\mathcal{P}\}_t \gets$ \textsc{DefinePlanerRegion}(LP) \Comment{$\{\mathcal{P}\}_t$ are planer regions at time $t$.} 
	\State $\{\mathcal{R}\}_t \gets$ \textsc{DefinePPGRegion}($P_t$) \Comment{$\{\mathcal{R}\}_t$  are the PPG ROIs for $t^{th}$ frame}
	\State $GF_0 \gets$ \textsc{GoodFeatures}($V(x,y,t),P_t$) \Comment{good features inside each region}
\Else
	\State $GF_1 \gets$ \textsc{TrackerKLT}($V(x,y,t)$,$V(x,y,t-1)$,$GF_0$)
	\State $AM_t \gets $ \textsc{AffineRANSAC}($GF_1,GF_0$)
	\Statex \Comment{$AM$ denote the affine model parameter computed separately for each planar region}
	\State $\{\mathcal{P}\}_t \gets \{\mathcal{P}\}_{t-1} \times AM_t$ 
	\State $\{\mathcal{R}\}_t \gets \{\mathcal{R}\}_{t-1} \times AM^P_t$
	\Statex \Comment{Each PPG region in set $\{\mathcal{R}\}$ tracked using encompassing planar region motion model}
	\State $GF_0 \gets GF_1$ 
\EndIf

\State $t \gets t+1$
\EndWhile
\State \textbf{return} $\{\mathcal{R}\}_t$ \Comment{Return tracked PPG regions $\{\mathcal{R}\}_t$ locations}

\EndProcedure
\end{algorithmic}
\end{algorithm}

\subsection{Implementation details} 
The ROIs inside the face region, $\mathcal{R}_i$, are chosen to be of the size of $20\times20$ pixel block. We have experimented with other sizes such as $10 \times 10$, $40\times40$~px, but have found $20\times20$ to be the best choice in terms of overall performance for our data-set. Note that the best choice of ROI size may depend on camera resolution and distance of subject from the camera ($0.5$~m for our data collection). We reinitialize the deformable face model and restart the KLT feature tracker every $T=5-10$~sec. KLT feature tracker is restarted because the local feature based motion tracker accumulates error over time. We have found that restarting KLT feature tracker every $T=5-10$~sec is sufficient for faithful facial region tracking during motions like reading, watching video or talking. Selecting a larger time interval $T$ helps in improving accuracy of goodness metric estimate, and smaller time interval helps in improving tracking performance. We balance this tradeoff by selecting $T=10$~sec when the person is relatively static, and choosing $T=5$~sec when there is motion. At the end of every $T=5-10$~sec (referred as an epoch from now on), we combine the PPG signal from all the tracked ROI using the \mrc\ algorithm. 

Within each epoch, we drop those KLT features that show more than $2$~px of forward-backward error. Further, we also reject those rigid regions $\mathcal{P}$ where there are fewer than $10$ KLT points left to track, as these regions might be undergoing  excessive motion or large illumination changes, and  including them will deteriorate the overall PPG estimate. For the RANSAC algorithm, we set the tolerance $\epsilon=2$~pixels, inlier fraction $\tau_{in}=0.7$, and maximum number of iteration $N=20$. If a rigid affine model cannot be found by RANSAC within this tolerance limit, we reject that region.  

Next, within each epoch,  we filter the PPG signal obtained from well tracked ROI ($\mathcal{R}_i$) using a zero-phase non-causal forward-backward bandpass filter, \filterBand. We then reject those ROI where the signal amplitude crosses the amplitude threshold $A_{th} = 8$. Camera-based PPG signals are really weak, and hence they hardly cross $\pm 2$ value in the units of $8$-bit camera pixel intensity. Any large change in intensity is mostly due to change in illumination or large motion artifact as discussed earlier. We then compute a coarse estimate of the pulse rate $PR_c$, by combining the PPG signals from all the remaining ROIs. This coarse pulse rate estimate can be erroneous at times, and so we keep track of the history of pulse rates over last $4$ epochs. If the current estimate of $PR_c$ is off by more than $\pm24~$bpm, then we replace the current estimate with the median of the last four estimate. 

We then compute the goodness metric $G_i$ for all the remaining ROI (all regions rejected prior to this stage are given a weight $G_i=0.0$ for the current epoch) and combine the PPG signals using the weighted average (Equation~\eqref{eg:PPG_estimate}).  We recompute the goodness metric $G_i$ for each ROI after every epoch.

\section{Dataset and Performance Evaluation}

\subsection{Prior methods for comparison}

For comparison, we implemented known past methods for camera-based PPG estimation. The general steps taken are: (i)  select the face regions using Voila Jones face detector as described in \cite{poh_non-contact_2010}, (ii) extract PPG signal by first computing the spatial average of the pixel intensity within the selected face region, and then filtering (detrending) the estimate  \cite{mcduff_improvements_2014,poh_non-contact_2010,verkruysse_remote_2008}, (iii) to compensate for motion, compute the 2D shift in face between consecutive frames as described in \cite{sun_motion-compensated_2011} and extracted the PPG from the tracked face region. In this sense, we have used a combination of known methods for single channel (green) camera-based PPG estimation algorithm for comparison with our \algo\ algorithm. We label this combination as \emph{\oldMethod}\ from now onwards. 

Another set of work in camera-based PPG estimation involves decomposing different camera channels (e.g. red, green, blue) into independent source signals using independent component analysis (ICA), and extracting the desired PPG  signal from one of these independent sources \cite{poh_non-contact_2010, poh_advancements_2011,mcduff_improvements_2014}. Our proposed \algo\ algorithm provides improvement in camera-based PPG estimate by spatially combining PPG signals from different regions of the face, and by improving upon the tracking algorithm, and uses only a single channel (green) of the camera. On the other hand, ICA based methods utilizes multiple camera channels (e.g. red, green, and blue \cite{poh_non-contact_2010} or cyan, orange, and green \cite{mcduff_improvements_2014}) to improve the performance of camera-based PPG estimate by separating independent sources. Thus, these two methods are characteristically different, and we will summarize the performance improvement provided by ICA-based method and by \algo.

\subsection{Experimental setup}

For all single channel video recording in this study, we used Flea 3\textsuperscript{\textregistered} USB 3.0 FL-U3-13E4M-C monochrome camera operated at $30$ frames per second, with a resolution of $1280\times1024$, and $8$~bits per pixel. We added a $40$~nm full-width half-max (FWHM) Green filter (FB550-40 from Thor labs \textsuperscript{\textregistered}) in front of the monochrome camera. We selected green filter since the absorption spectra of Hb and HbO2 peaks in this wavelength region. Moreover, all commercial color camera have highest number of pixel having green filter (Bayer pattern). For color video recording, we used Flea 3\textsuperscript{\textregistered} USB 3.0 FL-U3-13E4C-C color camera operated at $30$ frames per second with RGB Bayer pattern having a total resolution of $1280\times1024$, and $8$~bits per pixel.

We used Texas Instruments AFE4490SPO2EVM \pulseox\ module to record contact-based PPG signal for comparison. It operates at a sampling rate of $500$ Hz. The distance between camera and subject was $0.5$m. Both the camera system and the pulse-oximeter were started simultaneously, and the data is recorded in a PC workstation. All processing is done using a custom built software written in MATLAB.  All experiments (except the lighting experiment)  is conducted under ambient (fluorescent) lighting at 500 lux brightness.

\subsection{Dataset}
The main goal of the experiments reported here is to characterize and quantify the performance of the two  components of our proposed \algo\ algorithm ---  \mrc\ algorithm and \motion\ algorithm, and compare them with \oldMethod. We evaluate the performance by varying three main parameters of interest: (i) skin tone of people, (ii) motion, (iii) ambient light intensity, and quantify their effect on PPG estimation accuracy. We use the same raw video feed to evaluate our \algo\ algorithm in comparison to \oldMethod, and so performance improvements reported here are due to the proposed algorithm, and is independent of any specific hardware or camera choices we made for the evaluation.

All the experiments done in this research were approved by the Rice University institute review board (Protocol number: $14-145$E, Approval Date: $3/04/2014$). For the first experiment, we collected single channel (green) video data from $12$ subjects ($7$ male, $5$ female) with different skin tones (from light, pale white to dark brown/black). For this experiment, subjects were asked to face the camera and be static for a duration of $40$ seconds (involuntary motions were not restricted).

For the second set of experiments, we collected (single green channel) video under three natural motion scenarios --- (i) reading on computer, (ii) watching video, (iii) talking. The motion scenarios are  representative of a general class of motion exhibit by users of tablet, phone, or laptops while facing the screen. Reading scenario involves lateral movement of the head while reading text on screen. Watching video also involves intermittent facial expression such as smiling, getting amazed, sad etc, apart from lateral movement of the head. Talking scenario involves a lot of non-rigid movement around the jaw and the cheeks, and thus are ideal to evaluate system perform in harsh scenarios. For each of the motion scenarios, we collected $80$ seconds video recordings for $5$ subjects having different skin tones, along with their stationary video recording for baseline comparison. 

For the third set of experiments, we varied the illumination level from $50$ lux upto $650$ lux (ambient light is around $400-500$ lux) and recorded the single channel (green) video  for a duration of $40$ seconds under each lighting conditions for two subjects having pale-white and brown skin tones.

For comparison with ICA-based method, we collected two sets of data: (i) static dataset comprising of color video (red, green, blue) of $4$ subjects of varying skin tones (only non-Caucasian) for a duration of $60$ seconds at $500$~lux illumination, (ii) talking dataset comprising of color video of $4$ subjects of varying skin tone (non-Caucasian), with $3$ subjects having $500$~lux illumination, and $1$ subject having $300$~lux illumination. This set is deliberately chosen to be extremely harsh (lower light, non-Caucasian skin tones, and large motion during talking) to highlight scenario where current known algorithms (including \algo) fails, and provide dataset where future algorithms can improve. 

The dataset will be released in public and other researchers can access it at \url{http://www.ece.rice.edu/\texttildelow mk28/distancePPG/dataset/}.

\subsection{Performance metric}

As a waveform estimation algorithm, \algo\  provides an estimate of the ground truth PPG waveform $p(t)$ using video of a person's face. Thus, we use signal-to-noise ratio (SNR) of the estimate $\hat{p}(t)$ to quantify performance for comparison. The ground truth signal, $p(t)$, is recorded using a \pulseox\ attached to subject's ear. We chose earlobe instead of finger probe because of its proximity to the face region. Further, we also evaluate the performance based on the accuracy of physiological parameter like pulse rate and pulse rate variability  (i.e. beat-to-beat changes in pulse interval) that can be extracted from PPG waveform. We report the mean error rates in estimating PR and PRV using our \algo\ algorithm and using \oldMethod\ under different experimental scenarios. 

\subsubsection{SNR Definition}\label{sec:SNR}
To define the SNR of the estimated PPG waveform, we used standard PPG signal obtained from \pulseox\ connected to a person's earlobe as our ground truth signal.  The PPG waveform obtained from a contact \pulseox\ would also exhibit error due to motion and ambient light artifacts, and in that sense our choice of ground truth is a best effort choice. Nonetheless, the noise present in PPG waveform acquired using a contact \pulseox\ is orders of magnitude smaller than that obtained using a camera-based system, so our estimate of SNR would still be reasonably accurate.

The amplitude of the PPG signal recorded by \pulseox\ is unrelated to the amplitude of the camera-based PPG signal as both systems have completely independent sensor architecture and analog gain stage. So, here we have developed a definition of SNR which is independent of exact amplitude of these waveform. Our SNR metric captures how similar camera-based PPG signal is to the \pulseox\ based ground-truth PPG signal. 

Let $z(t)$ denote the PPG acquired using \pulseox. Let $k(t)$ denote the PPG estimated from a camera-based system ($k(t) = \hat{p}(t)$ from previous section). 
\begin{eqnarray}
k(t) & = & A_1p(t) + n_k(t) \\ 
z(t) & = & A_2p(t) + n_z(t)
\end{eqnarray}
Here, $n_k(t)$ is the noise in the PPG signal acquired from camera. Apart from the quantization noise, $n_k(t)$  also includes uncompensated motion artifacts. The noise present in the \pulseox\ system is denoted as $n_z(t)$. Let us assume that all signals are defined in the time interval $[0,T]$ and we use $L_2$ inner product and norms for all definitions.  

The noise $n_k(t)$ would be uncorrelated to $n_z(t)$ as both acquisition systems are unrelated, and both noise is uncorrelated to the underlying zero mean PPG signal $p(t)$. If we assume that all signals follow ergodicity and consider integration over large time window ($T$ is large), then
\begin{eqnarray}
\langle n_k(t),n_z(t)\rangle & \approx & 0 \\ 
\langle n_k(t) ~\text{or}~ n_z(t),p(t)\rangle & \approx & 0.
\end{eqnarray}

and further, we can write
\begin{equation}
\frac{\langle k(t),z(t)\rangle}{\langle z(t),z(t)\rangle} \approx \frac{A_1A_2\cdot\big\|p(t)\big\|^2}{A_2{}^2\cdot\big\|p(t)\big\|^2 + \big\|n_z(t)\big\|^2 }
\end{equation}

Since the signal quality of PPG derived from \pulseox\ is reasonably good, we can assume that noise power $\big\|n_z(t)\big\|^2$ is much smaller than signal power $\big\|p(t)\big\|^2$ in the denominator. Thus, $\frac{A_1}{A_2}$ can be approximated as 
\begin{equation}
\frac{A_1}{A_2} \approx \frac{\langle k(t),z(t)\rangle}{\langle z(t),z(t)\rangle}.
\end{equation}
Further, if we assume that $\big\|n_z(t)\big\|^2 \ll \big\|n_k(t)\big\|^2  $, we can estimate the noise present in $k(t)$ as
\begin{equation}
n_k(t) \equiv k(t) - \frac{\langle k(t),z(t)\rangle}{\langle z(t),z(t)\rangle}\cdot z(t)
\end{equation}
and, underlying signal $s_k(t)$ can be approximated to 
\begin{equation}
s_k(t) \equiv \frac{\langle k(t),z(t)\rangle}{\langle z(t),z(t)\rangle}\cdot z(t),
\end{equation}
which leads us to define signal to noise ration or SNR simply as 
\begin{equation}\label{eq:SNR}
\text{SNR} \equiv \frac{\big\|s_k(t)\big\|^2}{\big\|n_k(t)\big\|^2}.
\end{equation}

The SNR measure defined in (\ref{eq:SNR}) will be used in the following sections to compare performance of \algo\ algorithm and \oldMethod\ under various experimental scenarios. 
 
\subsubsection{Quantification of physiological parameters}
To estimate the pulse rate, we compute the spectrum of the PPG signal using FFT algorithm over every $10$~sec window (hamming window) with $5$~sec overlap. The pulse rate is estimated as the frequency that correspond to the highest power in the estimated spectrum (PR $=60\cdot f_{PR}$~bpm). Similarly, we obtain the reference pulse rate from synchronously acquired contact PPG. 

To estimate the PRV or inter-beat interval (IBI), we first interpolate the camera-based PPG signal with a cubic spline function to a sampling rate of $500$~Hz (same as the sampling rate for \pulseox). We then detect the peaks in the interpolated PPG signal using a custom algorithm for  minima-detection. To avoid inclusion of artifacts due to noise or motion, we use the estimated pulse rate to reject peaks within $0.5/f_{PR}$ time difference. The time interval between consecutive peaks are the PRV (or IBI).  Same algorithm is used to find the reference PRV from synchronously acquired contact PPG. 

Due to the noisy estimate of PPG signal for darker skin tones and/or under large motion (e.g. during talking), some pulse beats (peaks)  cannot be detected in camera PPG signal. We determine the number of missing peaks by comparing peak location in camera PPG with the reference \pulseox\ PPG.  Thus, we also report the percentage of missing  peaks in the camera-based PPG signal as a performance metric associated with PRV estimation, apart from the RMSE of PRV.

\newcommand{\bottomLeftplotPPG}[1] {
	\begin{axis}[
			name=old,
			width=2.5in,
			height=1in,
			xmin=-1,
			xmax=40,
			xtick={\empty},
			ymin=-1.5,
			ymax=1.5,
			ytick={-1,1},
			axis x line = none,
			axis y line = left,
			y axis line style=-,
			legend columns=3,
			legend style={at={(1.2,+1.15)},anchor=south},
			]
			\addplot [color=orange,solid, thin] table[x=time,y=old]{#1};
			\addlegendentry{\oldMethod\ (top)}
			\addplot [color=violet, thin] coordinates{(-1,0)};
			\addlegendentry{\algo\ (middle)}
			\addplot [color=black, thin] coordinates{(-1,0)};
			\addlegendentry{ground truth (bottom)}
			
		\end{axis} 
		\begin{axis} [
			name=mrc,
			at={($(old.south)-(0,0.2cm)$)},
			width=2.5in,
			height=1in,
			anchor=north,
			xmin=-1,
			xmax=40,
			xtick={\empty},
			ymin=-1.5,
			ymax=1.5,
			ytick={-1,1},
			ylabel=Camera PPG Scale,
			axis x line = none,
			axis y line = left,
			y axis line style=-,	
			]
			\addplot [color=violet,solid,  thin] table[x=time,y=mrc]{#1};
		\end{axis} 
		\begin{axis} [
			name=signal,
			at={($(mrc.south)-(0,0.2cm)$)},
			width=2.5in,
			height=1in,
			anchor=north,
			xmin=-1,
			xmax=40,
			axis x line = bottom,
			axis y line = left,
			y axis line style=-,
			ymin=-1.5,
			ymax=1.5,
			ytick={-1,1},
			xlabel=Time(s),]
			\addplot [color=black, thin] table[x=time,y=signal]{#1};
		\end{axis}

}

\newcommand{\bottomRightPlotPPG}[1] {
	\begin{axis}[
			name=old,
			width=2.5in,
			height=1in,
			xmin=-1,
			xmax=40,
			xtick={\empty},
			ymin=-1.5,
			ymax=1.5,
			ytick={-1,1},
			axis x line = none,
			axis y line = left,
			y axis line style=-,
			]
			\addplot [color=orange,solid, thin] table[x=time,y=old]{#1};
		\end{axis} 
		\begin{axis} [
			name=mrc,
			at={($(old.south)-(0,0.2cm)$)},
			width=2.5in,
			height=1in,
			anchor=north,
			xmin=-1,
			xmax=40,
			xtick={\empty},
			ymin=-1.5,
			ymax=1.5,
			ytick={-1,1},
			axis x line = none,
			axis y line = left,
			y axis line style=-,	
			]
			\addplot [color=violet,solid,  thin] table[x=time,y=mrc]{#1};
		\end{axis} 
		\begin{axis} [
			name=signal,
			at={($(mrc.south)-(0,0.2cm)$)},
			width=2.5in,
			height=1in,
			anchor=north,
			xmin=-1,
			xmax=40,
			axis x line = bottom,
			axis y line = left,
			y axis line style=-,
			ymin=-1.5,
			ymax=1.5,
			ytick={-1,1},
			xlabel=Time(s)
			]
			\addplot [color=black,solid,  thin] table[x=time,y=signal]{#1};
		\end{axis}

}

\newcommand{\topRightplotPPG}[1] {
	\begin{axis}[
			name=old,
			width=2.5in,
			height=1in,
			xmin=-1,
			xmax=40,
			xtick={\empty},
			ymin=-1.5,
			ymax=1.5,
			ytick={-1,1},
			axis x line = none,
			axis y line = left,
			y axis line style=-,
			]
			\addplot [color=orange,solid, thin] table[x=time,y=old]{#1};
		\end{axis} 
		\begin{axis} [
			name=mrc,
			at={($(old.south)-(0,0.2cm)$)},
			width=2.5in,
			height=1in,
			anchor=north,
			xmin=-1,
			xmax=40,
			xtick={\empty},
			ymin=-1.5,
			ymax=1.5,
			ytick={-1,1},
			ylabel=Camera PPG Scale,
			axis x line = none,
			axis y line = left,
			y axis line style=-,	
			]
			\addplot [color=violet,solid,  thin] table[x=time,y=mrc]{#1};
		\end{axis} 
		\begin{axis} [
			name=signal,
			at={($(mrc.south)-(0,0.2cm)$)},
			width=2.5in,
			height=1in,
			anchor=north,
			xmin=-1,
			xmax=40,
			axis x line = bottom,
			axis y line = left,
			y axis line style=-,
			ymin=-1.5,
			ymax=1.5,
			ytick={-1,1},
			xlabel=Time(s)
			]
			\addplot [color=black,solid,  thin] table[x=time,y=signal]{#1};
		\end{axis}

}

\newcommand{\topRightPlotPPGMotion}[2] {
	\begin{axis}[
			name=old,
			width=2.5in,
			height=1in,
			xmin=-1,
			xmax=40,
			xtick={\empty},
			ymin=-1.5,
			ymax=1.5,
			ytick={-1,1},
			axis x line = none,
			axis y line = left,
			y axis line style=-,
			]
			\addplot [color=orange,thin] table[x=time,y=old]{#1};
		\end{axis} 
		\begin{axis} [
			name=mrc,
			at={($(old.south)-(0,0.2cm)$)},
			width=2.5in,
			height=1in,
			anchor=north,
			xmin=-1,
			xmax=40,
			ylabel=Camera PPG Scale,
			xtick={\empty},
			ymin=-1.5,
			ymax=1.5,
			ytick={-1,1},				
			axis x line = none,
			axis y line = left,
			y axis line style=-,	
			]
			\addplot [color=violet,thin] table[x=time,y=mrc]{#1};
		\end{axis} 
		\begin{axis} [
			name=signal,
			at={($(mrc.south)-(0,0.2cm)$)},
			width=2.5in,
			height=1in,
			anchor=north,
			xmin=-1,
			xmax=40,
			axis x line = bottom,
			axis y line = left,
			y axis line style=-,
			xtick={\empty},
			ymin=-1.5,
			ymax=1.5,
			ytick={-1,1},]
			\addplot [color=black,thin] table[x=time,y=signal]{#1};
		\end{axis}
		
		\begin{axis} [
			name=motion,
			at={($(signal.south)-(0,0.3cm)$)},
			width=2.5in,
			height=1.0in,
			anchor=north,
			xmin=-1,
			xmax=40,
			ylabel=motion (px),
			axis x line = bottom,
			axis y line = left,
			y axis line style=-,
			xlabel={Time (sec)},
			ymin=0,
			ymax=10,
			ytick={0,5},]
			\addplot+[color=black, const plot mark mid, mark size=0.5pt] table[x=time,y=motion]{#2};
		\end{axis}

}

	\newcommand{\bottomLeftPlotPPGMotion}[2] {
	\begin{axis}[
			name=old,
			width=2.5in,
			height=1in,
			xmin=-1,
			xmax=40,
			xtick={\empty},
			ymin=-1.5,
			ymax=1.5,
			ytick={-1,1},
			axis x line = none,
			axis y line = left,
			y axis line style=-,
			legend columns=4,
			legend style={at={(1.2,+1.15)},anchor=south},
			]
			\addplot [color=orange,solid, thin] table[x=time,y=old]{#1};
			\addlegendentry{\oldMethod}
			\addplot [color=violet, thin] coordinates{(-1,0)};
			\addlegendentry{\algo}
			\addplot [color=black, thin] coordinates{(-1,0)};
			\addlegendentry{ground truth}
			\addplot+[color=black, const plot mark mid, mark size=0.5pt] coordinates{(-1,0)};
			\addlegendentry{avg motion per frame(px)}
		\end{axis} 
		\begin{axis} [
			name=mrc,
			at={($(old.south)-(0,0.2cm)$)},
			width=2.5in,
			height=1in,
			anchor=north,
			xmin=-1,
			xmax=40,
			xtick={\empty},
			ymin=-1.5,
			ylabel=Camera PPG Scale,
			ymax=1.5,
			ytick={-1,1},				
			axis x line = none,
			axis y line = left,
			y axis line style=-,	
			]
			\addplot [color=violet,thin] table[x=time,y=mrc]{#1};
		\end{axis} 
		\begin{axis} [
			name=signal,
			at={($(mrc.south)-(0,0.2cm)$)},
			width=2.5in,
			height=1in,
			anchor=north,
			xmin=-1,
			xmax=40,
			axis x line = bottom,
			axis y line = left,
			y axis line style=-,
			xtick={\empty},
			ymin=-1.5,
			ymax=1.5,
			ytick={-1,1},]
			\addplot [color=black,thin] table[x=time,y=signal]{#1};
		\end{axis}
		
		\begin{axis} [
			name=motion,
			at={($(signal.south)-(0,0.3cm)$)},
			width=2.5in,
			height=1.0in,
			anchor=north,
			xmin=-1,
			xmax=40,
			ylabel=motion (px),
			axis x line = bottom,
			axis y line = left,
			y axis line style=-,
			xlabel={Time (sec)},
			ymin=0,
			ymax=12,
			ytick={0,5},]
			\addplot+[color=black, const plot mark mid, mark size=0.5pt] table[x=time,y=motion]{#2};
		\end{axis}

}

\newcommand{\bottomRightPlotPPGMotion}[2] {
	\begin{axis}[
			name=old,
			width=2.5in,
			height=1in,
			xmin=-1,
			xmax=40,
			xtick={\empty},
			ymin=-1.5,
			ymax=1.5,
			ytick={-1,1},
			axis x line = none,
			axis y line = left,
			y axis line style=-,
			]
			\addplot [color=orange,solid,thin] table[x=time,y=old]{#1};
		\end{axis} 
		\begin{axis} [
			name=mrc,
			at={($(old.south)-(0,0.2cm)$)},
			width=2.5in,
			height=1in,
			anchor=north,
			xmin=-1,
			xmax=40,
			xtick={\empty},
			ymin=-1.5,
			ymax=1.5,
			ytick={-1,1},				
			axis x line = none,
			axis y line = left,
			y axis line style=-,	
			]
			\addplot [color=violet,solid] table[x=time,y=mrc]{#1};
		\end{axis} 
		\begin{axis} [
			name=signal,
			at={($(mrc.south)-(0,0.2cm)$)},
			width=2.5in,
			height=1in,
			anchor=north,
			xmin=-1,
			xmax=40,
			axis x line = bottom,
			axis y line = left,
			y axis line style=-,
			xtick={\empty},
			ymin=-1.5,
			ymax=1.5,
			ytick={-1,1},]
			\addplot [color=black,solid] table[x=time,y=signal]{#1};
		\end{axis}
		
		\begin{axis} [
			name=motion,
			at={($(signal.south)-(0,0.3cm)$)},
			width=2.5in,
			height=1.0in,
			anchor=north,
			xmin=-1,
			xmax=40,
			axis x line = bottom,
			axis y line = left,
			y axis line style=-,
			xlabel={Time (sec)},
			ymin=0,
			ymax=8,
			ytick={0,5},]
			\addplot+[color=black, const plot mark mid, mark size=0.5pt] table[x=time,y=motion]{#2};
		\end{axis}

}

\section {Results}
\subsection{Performance for different skin tones}

\begin{figure}
\centering

	\begin{subfigure}[t]{0.48\textwidth}
	\centering
	\begin{tikzpicture}
		 \begin{axis}[
		    	ybar=5pt,
		    	width=2.5in,
		    	height =2in,
		    	axis x line = bottom,
		    	axis y line = left,
		    	ylabel=SNR(dB), 
		    	ymin=0,
		    	ymax=12,
		    	xmin=0.5,
		    	xmax=3.5,
			   	xtick={1,2,3},
			   	nodes near coords,
			   	nodes near coords align={vertical},
		      	xticklabels={fair, olive,brown},
		      	yticklabel style={/pgf/number format/fixed},
			   ylabel=SNR (dB),
			   x tick label style={rotate=45,anchor=east},
		      legend style={at={(0.5,-0.25)},anchor=north},
		      legend columns=2,
		      ]
		  \addplot[
		  black,
		  fill=orange!30!white,
		  mark=none,
		  postaction={pattern=horizontal lines},
		  ] table[x=IDs,y=OLD_SNR] {result/SNR_static.txt};
	  	
	  	\addplot[
	  	black,
	  	fill=purple!30!white!,
	  	mark=none,
	  	postaction={pattern=north east lines},
	  	] table[x=IDs,y=MRC_SNR] {result/SNR_static.txt};
	
	  \legend{\oldMethod,\algo}
	  
	  \end{axis}
	\end{tikzpicture}
	\caption{SNR gain due to \algo}
	\label{fig:static_SNR}
	\end{subfigure}	\begin{subfigure}[t]{0.48\textwidth}
	\centering 
	\begin{tikzpicture}
	\topRightplotPPG{result/raw_static_fair.txt}
	\end{tikzpicture}

	\caption{Fair skin tone}
	\label{fig:static_waveform_fair}
	\end{subfigure}		
	\begin{subfigure}[t]{0.48\textwidth}
	\centering
	\begin{tikzpicture}[baseline]
	
	\bottomLeftplotPPG{result/raw_static_medium.txt}

	\end{tikzpicture}
	\caption{Olive skin tone}
	\label{fig:static_waveform_medium}
	\end{subfigure}	\begin{subfigure}[t]{0.48\textwidth}
	\centering
	\begin{tikzpicture}[baseline]
	\bottomRightPlotPPG{result/raw_static_brown.txt}

	\end{tikzpicture}
	
	\caption{Brown skin tone}
	\label{fig:static_waveform_brown}
	\end{subfigure}

	\caption{\algo\ performance comparison for different skin tones:  Plot (a) shows SNR improvement due to \algo\ for subjects clubbed into three categories: fair/light ($4$), olive ($4$), brown ($4$). The SNR gain due to \algo\ is more for darker skin tones. Plot (b)-(d) compares typical camera PPG waveform estimated using \oldMethod\ (top) and \algo\ (middle) with the ground truth \pulseox\ signal (bottom).}
	\label{fig:static_SNR_waveform}
\end{figure}
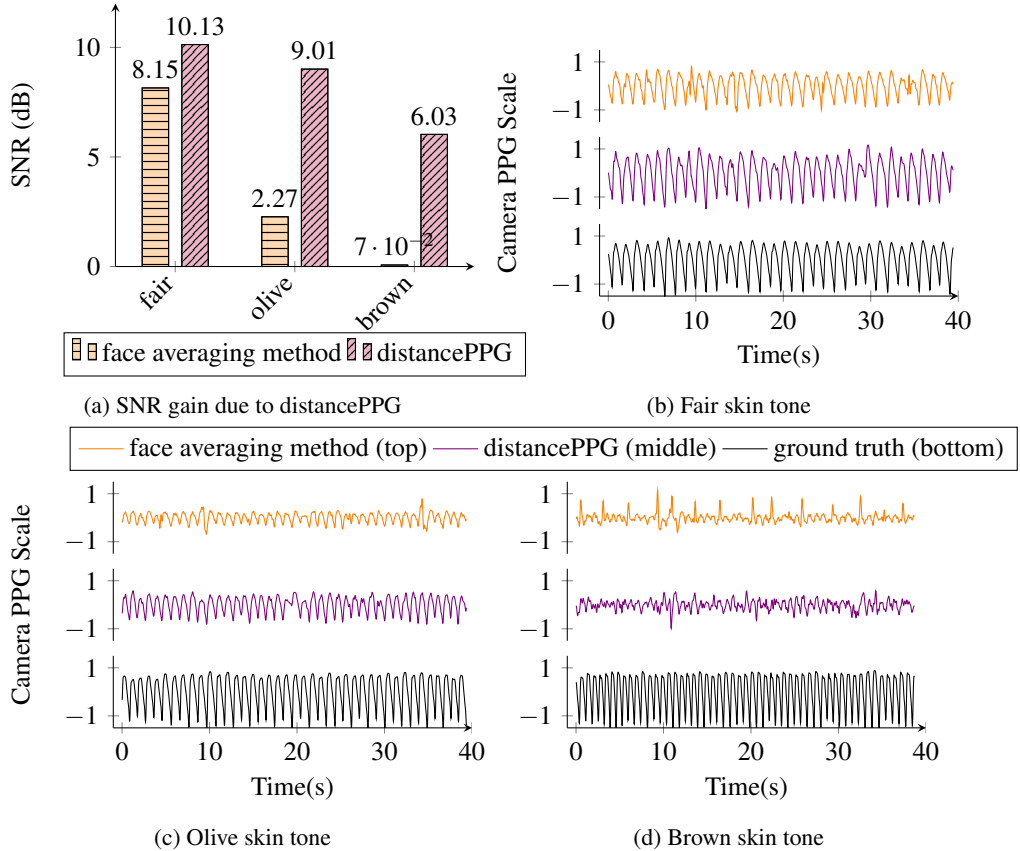

Figure~\ref{fig:static_SNR_waveform} compares the SNR and the waveform of PPG obtained using \algo\ and \oldMethod\ for people having different skin tone categories --- (i) light/fair ($4$ people), (ii) medium/olive ($4$ people), and (iii) brown/dark ($4$ people). As seen in Figure~\ref{fig:static_SNR}, on an average \algo\ provides $\DSNRStatic$~dB of SNR improvement for all skin tones compared to \oldMethod. Also, the gain in SNR is more for darker skin tones (around $\DSNRDark$~dB). Further, Figure~\ref{fig:static_waveform_fair}-\ref{fig:static_waveform_brown} shows an example $40$~s plots of camera-based PPG estimate using both \algo\ and \oldMethod\ for the three skin tone categories.  When compared with ground truth signal, it is evident that camera-based PPG signal estimated using \algo\ is more similar to the ground truth \pulseox\ signal.

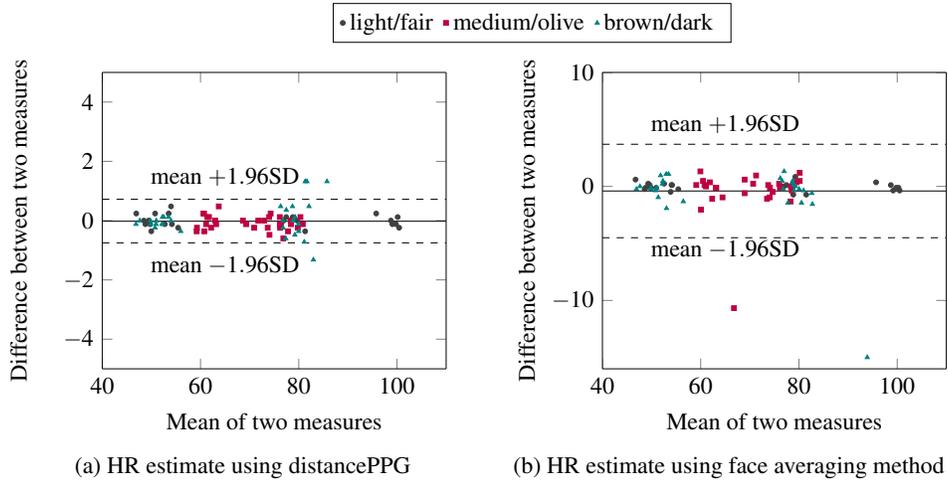
\begin{figure}[h]
\centering

	\begin{subfigure}[t]{0.48\textwidth}
	\centering 
		\begin{tikzpicture}[scale=0.9,baseline]
		\begin{axis}[%
			at={(0in,0in)},
			width=2in,
			scale only axis,
			separate axis lines,
			every outer x axis line/.append style={black},
			every x tick label/.append style={font=\color{black}},
			xmin=40,
			xmax=110,
			xlabel={Mean of two measures},
			every outer y axis line/.append style={black},
			every y tick label/.append style={font=\color{black}},
			ymin=-5,
			ymax=5,
			ylabel={Difference between two measures},
			legend style={at={(1.25,1.1)},anchor=south, draw=black},
			legend columns=3,	]
		\addplot [color=darkgray,mark size=1pt,only marks,mark=*,mark options={solid}] table[x=mean,y=diff]{result/bland_static_data_max_fair.txt};
		\addlegendentry{light/fair \qquad };
		
		\addplot [color=purple,mark size=1pt,only marks,mark=square*,mark options={solid}] table[x=mean,y=diff]{result/bland_static_data_max_medium.txt};
		\addlegendentry{medium/olive \qquad };
		
		\addplot [color=teal,mark size=1pt,only marks,mark=triangle*,mark options={solid}] table[x=mean,y=diff]{result/bland_static_data_max_brown.txt};
		\addlegendentry{brown/dark \qquad};
		
		\addplot [color=black,solid,forget plot] coordinates {(40,\DmrcHRMD) (110,\DmrcHRMD)};
		\addplot [color=black,dashed,forget plot] coordinates {(40,\DmrcHRMDminusSD) (110,\DmrcHRMDminusSD)};
		  
		\node at (axis cs: 65,-1.5) {mean $- 1.96$SD };
		\addplot [color=black,dashed,forget plot]  coordinates {(40,\DmrcHRMDplusSD) (110,\DmrcHRMDplusSD)};
		 
		\node at (axis cs: 65,+1.5) {mean $+ 1.96$SD };

		\end{axis}

		\end{tikzpicture}
		\caption{HR estimate using \algo}
	\end{subfigure}	\begin{subfigure}[t]{0.48\textwidth}
	\centering
		\begin{tikzpicture}[scale=0.9,baseline]
		\begin{axis}[%
			at={(0in,0in)},
			width=2in,
			scale only axis,
			separate axis lines,
			every outer x axis line/.append style={black},
			every x tick label/.append style={font=\color{black}},
			xmin=40,
			xmax=110,
			xlabel={Mean of two measures},
			every outer y axis line/.append style={black},
			every y tick label/.append style={font=\color{black}},
			ymin=-16,
			ymax=10,
			ylabel={Difference between two measures},
			ylabel near ticks, 
			]
		\addplot [color=darkgray,mark size=1pt,only marks,mark=*,mark options={solid}] table[x=mean,y=diff]{result/bland_static_data_old_fair.txt};
		
		\addplot [color=purple,mark size=1pt,only marks,mark=square*,mark options={solid}] table[x=mean,y=diff]{result/bland_static_data_old_medium.txt};
		
		\addplot [color=teal,mark size=1pt,only marks,mark=triangle*,mark options={solid}] table[x=mean,y=diff]{result/bland_static_data_old_brown.txt};
		
		\addplot [color=black,solid,forget plot] coordinates {(40,\DoldHRMD) (110,\DoldHRMD)};
		\addplot [color=black,dashed,forget plot] coordinates {(40,\DoldHRMDminusSD) (110,\DoldHRMDminusSD)};
			  
		\node at (axis cs: 65,-5.5) {mean $- 1.96$SD };
		\addplot [color=black,dashed,forget plot]  coordinates {(40,\DoldHRMDplusSD) (110,\DoldHRMDplusSD)};
				 
		\node at (axis cs: 65,+5.5) {mean $+ 1.96$SD };

		\end{axis}
		\end{tikzpicture}
		\caption{HR estimate using \oldMethod}
		
	\end{subfigure}
	\caption{Bland-Altman plot: comparison of PR derived from camera PPG and from ground truth \pulseox\  for different skin tones, $12$ subjects having fair ($4$), olive ($4$) and brown ($4$) skin tones categories. Note that different y-axis scales are used for \algo\ and \oldMethod\ to accommodate all the points. }
	\label{fig:stationary_bland}

\end{figure}

The improvement in SNR for various  skin tones reduces the error in PR estimation. Figure~\ref{fig:stationary_bland} shows the agreement between camera derived PR and ground truth \pulseox\ derived PR from $12$ subjects of different skin tones using Bland-Altman plot. For PPG estimated using \oldMethod, the mean bias (average difference between ground truth \pulseox\ derived PR and camera-based PPG derived PR) $\bar{d}$ = $\DoldHRMD$ bpm with $95\%$ limit of agreement (mean bias $\pm 1.96$ standard deviation of the difference) being $\DoldHRMDminusSD$ to $\DoldHRMDplusSD$ bpm. Using the \algo\ reduces the error in PR estimate to $\bar{d} =\DmrcHRMD$ bpm with $95\%$ limit of agreement being $\DmrcHRMDminusSD$ to $\DmrcHRMDplusSD$ bpm. For computing these statistics, we have not included occasional extreme outliers (error in PR $ \geq 30$ bpm) found in \oldMethod\ ($4$ outliers in $88$ sample points) as these outliers significantly skew the performance of \oldMethod\ and show it in bad light. \Algo\ algorithm did not exhibit any extreme outliers in the dataset for all skin tones and so we have reported statistics for all $88$ sample points.

\begin{figure}[h]
\centering

	\begin{subfigure}[t]{0.48\textwidth}
	\centering 
		\begin{tikzpicture}[baseline]

		    \begin{axis}[
		    	ybar=10pt,
		    	width=2.5in,
		    	height =2in,
		    	xtick={1,2,3},
		    	axis x line = bottom,
		    	axis y line = left, 
		    	ymin=0,		    			   
		    	ymax=108,
		    	xmin=0.5,
		    	xmax=3.5,
		    	nodes near coords,
			   	nodes near coords align={vertical},
		    	xticklabels={fair, olive,brown},		    	 x tick label style={rotate=45,anchor=east},
		        ylabel=RMSE in PRV estimate (ms),
		       legend style={at={(1.2,1.1)},anchor=south},
		       legend columns=2,]
		  \addplot[black, fill=orange!30!white,
		    	mark=none,
		    	postaction={pattern=north east lines},] table[x=IDs,y=OLD] {result/IBI_RMSE_static.txt};
		  \addplot[black, fill=purple!30!white,
		   		    		  mark=none,
		   		    		  postaction={pattern=horizontal lines},
		   		    		  ] table[x=IDs,y=MRV] {result/IBI_RMSE_static.txt};
		    \legend{\oldMethod \hspace{1cm},\algo}
		    \end{axis}
		\end{tikzpicture}
		\caption{RMSE of PRV}
	\end{subfigure} \begin{subfigure}[t]{0.48\textwidth}
	\centering												\begin{tikzpicture}[baseline]

		    \begin{axis}[
				ybar=10pt,
		    	width=2.5in,
		    	height =2in,
		    	xtick={1,2,3},
		    	axis x line = bottom,
		    	axis y line = left, 
		    	ymin=0,		    			   
		    	ymax=10,
		    	xmin=0.5,
		    	xmax=3.5,
		    	nodes near coords,
			   	nodes near coords align={vertical},
		    	ylabel= \% missing peaks, 
		    	xticklabels={fair, olive,brown}, 
		    	x tick label style={rotate=45,anchor=east},]
		   \addplot[black, fill=purple!30!white,
		   mark=none,
		   postaction={pattern=horizontal lines},
		   ] table[x=IDs,y=MRV] {result/missing_static.txt};
			\addplot[black, fill=orange!30!white,
		    	mark=none,
		    	postaction={pattern=north east lines},] table[x=IDs,y=OLD] {result/missing_static.txt};
		     \end{axis}
		\end{tikzpicture}
		\caption{Percentage of missing peaks}
	\end{subfigure} 
	\caption{Pulse rate variability (PRV) estimation performance for different skin tones: $12$ subjects having fair ($4$), olive ($4$) and brown ($4$) skin tones categories. (Left) Bar-plot  shows the RMSE in the timings of the peaks in the camera-based PPG waveform when compared with \pulseox\ based ground truth. (Right) bar-plot shows the percentage of missing peaks in camera-based PPG waveform. \Algo significantly reduces the percentage of missing peaks.}
	
	\label{fig:stationary_IBI}
\end{figure}
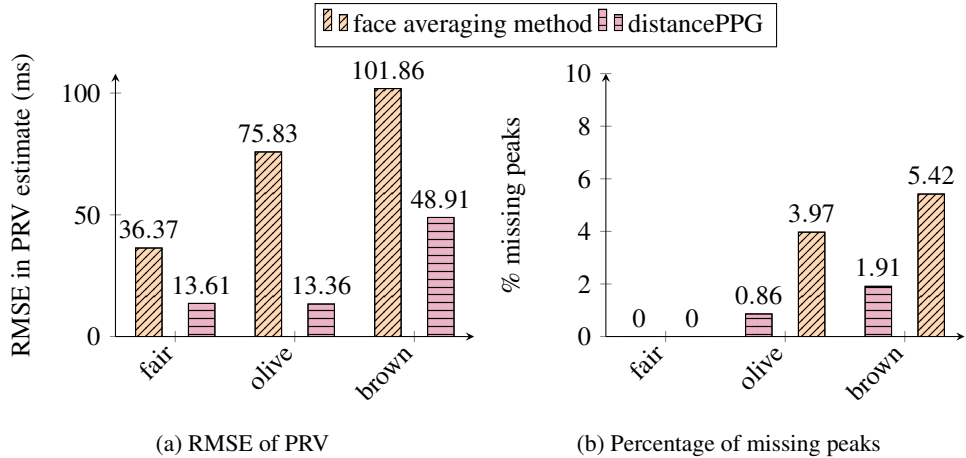

PPG signal estimated using \algo\ shows significant improvement in the root mean squared error (RMSE) of PRV estimate.  Figure~\ref{fig:stationary_IBI} shows the RMSE of PRV estimation and the corresponding percentage of missing peaks in camera-based PPG for the three skin tone categories. Using \algo, the RMSE in PRV for light and medium skin tone is less than $16$~ms. As $30$~fps camera is used for recording the video, one can not expect RMSE to be much lower than half the sampling interval ($16.6$~ms).

\subsection{Performance under various motion scenario}

\begin{figure}
\centering
	\begin{subfigure}[t]{0.48\textwidth}
	\centering 

	\begin{tikzpicture}
	
		 \begin{axis}[
			    	ybar=5pt,
			    	width=2.5in,
			    	height =2in,
			    	axis x line = bottom,
			    	axis y line = left,
			    	ylabel=SNR(dB), 
			    	ymin=-5,
			    	ymax=10,
			    	xmin=0.5,
			    	xmax=4.5,
				   	xtick={1,2,3,4},
				   	nodes near coords,
				   	nodes near coords align={vertical},
			      	xticklabels={stationary, reading,watching video, talking},
				   ylabel=SNR (dB),
				   x tick label style={rotate=45,anchor=east},
			      legend style={at={(0.5,1.10)},anchor=south},
			      legend columns=2,
			      ]
			  \addplot[
			  black,
			  fill=orange!30!white,
			  mark=none,
			  postaction={pattern=horizontal lines},
			  ] table[x=IDs,y=OLD_SNR] {result/SNR_motion.txt};
		  	
		  	\addplot[
		  	black,
		  	fill=purple!30!white!,
		  	mark=none,
		  	postaction={pattern=north east lines},
		  	] table[x=IDs,y=MRC_SNR] {result/SNR_motion.txt};
		
		  \legend{\oldMethod,\algo}
		  
		  \end{axis}

	
	\end{tikzpicture}
	\caption{SNR gain under motion due to \algo}
	\label{fig:motion_SNR}
	\end{subfigure}	\begin{subfigure}[t]{0.48\textwidth}
	\centering
	\begin{tikzpicture}
	\topRightPlotPPGMotion{result/raw_motion_read.txt}{result/motion_read.txt}

	\end{tikzpicture}
	\caption{reading scenario}
	\label{fig:motion_waveform_reading}
	\end{subfigure}
	\begin{subfigure}[t]{0.48\textwidth}
		\centering
		\begin{tikzpicture}[baseline]
		
		\bottomLeftPlotPPGMotion{result/raw_motion_watch.txt}{result/motion_watch.txt}
	
		\end{tikzpicture}
		\caption{Watching video }
		\label{fig:motion_waveform_watching}
		\end{subfigure}	\begin{subfigure}[t]{0.48\textwidth}
		\centering
		\begin{tikzpicture}[baseline]
		\bottomRightPlotPPGMotion{result/raw_motion_talk.txt}{result/motion_talk.txt}

		\end{tikzpicture}
		
		\caption{talking on skype}
		\label{fig:motion_waveform_talking}
		\end{subfigure}	
		
		\label{fig:motion_improvement}
		\caption{Performance of \algo under various motion scenario: Plot (a) shows the improvement in SNR due to \algo, (b-d) shows typical snippets of PPG waveform estimated using \oldMethod\ (top) and \algo\ (middle) during different types of motion. Comparing the camera PPG waveform with ground truth \pulseox\ (bottom) clearly shows the improvement due to \algo\ during small-medium motion ($\le4$~px per frame). Under large motion ($\geq5$~px per frame), \algo\ suffers due to uncompensated motion artifacts. Average motion magnitude captures only one aspect of motion and thus cannot always predict exact performance deterioration under various types of motion like turning, tilting etc.  }
	
\end{figure}

Figure~\ref{fig:motion_SNR} shows the SNR improvements for four scenarios --- (i) stationary, (ii) reading, (iii) watching video, and (iv) talking for $5$ subjects of different skin tones. \Algo\ provides on an average $\DSNRMotion$dB of SNR improvement under various motion scenario compared to \oldMethod. Further, Figure~\ref{fig:motion_waveform_reading}-\ref{fig:motion_waveform_talking}  shows typical $40$~s plots of camera-based PPG estimate using both \algo\ and \oldMethod\ for the three motion scenarios. When compared with ground truth signal, it is evident that \algo\ improves the estimate of PPG during small-medium motion ($\le4$~px per frame). Under large motion ($\geq5$~px per frame), \algo\ also suffers due to motion artifacts.

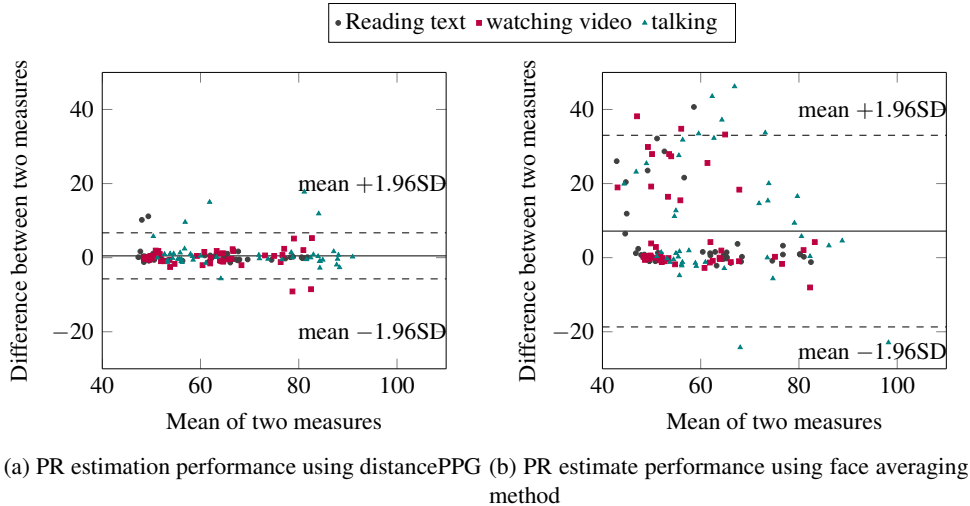
\begin{figure}[h]
\centering

	\begin{subfigure}[t]{0.48\textwidth}
	\centering 
		\begin{tikzpicture}[scale=0.9,baseline]
		\begin{axis}[%
			at={(0in,0in)},
			width=2in,
			scale only axis,
			separate axis lines,
			every outer x axis line/.append style={black},
			every x tick label/.append style={font=\color{black}},
			xmin=40,
			xmax=110,
			xlabel={Mean of two measures},
			every outer y axis line/.append style={black},
			every y tick label/.append style={font=\color{black}},
			ymin=-30,
			ymax=50,
			ylabel={Difference between two measures},
			legend style={at={(1.25,1.1)},anchor=south, draw=black},
			legend columns=3,	]
		\addplot [color=darkgray,mark size=1pt,only marks,mark=*,mark options={solid}] table[x=mean,y=diff]{result/bland_motion_data_max_read.txt};
		\addlegendentry{Reading text \qquad  };
		
		\addplot [color=purple,mark size=1pt,only marks,mark=square*,mark options={solid}] table[x=mean,y=diff]{result/bland_motion_data_max_watch.txt};
		\addlegendentry{watching video \qquad  };
		
		\addplot [color=teal,mark size=1pt,only marks,mark=triangle*,mark options={solid}] table[x=mean,y=diff]{result/bland_motion_data_max_talk.txt};
		\addlegendentry{talking \qquad };
		
		\addplot [color=black,solid,forget plot] coordinates {(40,\DmrcHRMDmotion) (110,\DmrcHRMDmotion)};
		\addplot [color=black,dashed,forget plot] coordinates {(40,\DmrcHRMDminusSDmotion) (110,\DmrcHRMDminusSDmotion)};
		  
		\node at (axis cs: 95,-20) {mean $- 1.96$SD };
		\addplot [color=black,dashed,forget plot]  coordinates {(40,\DmrcHRMDplusSDmotion) (110,\DmrcHRMDplusSDmotion)};
		 
		\node at (axis cs: 95,20) {mean $+ 1.96$SD };

		\end{axis}

		\end{tikzpicture}
		\caption{PR estimation performance using \algo}
	\end{subfigure}	\begin{subfigure}[t]{0.48\textwidth}
	\centering
		\begin{tikzpicture}[scale=0.9,baseline]
		\begin{axis}[%
			at={(0in,0in)},
			width=2in,
			scale only axis,
			separate axis lines,
			every outer x axis line/.append style={black},
			every x tick label/.append style={font=\color{black}},
			xmin=40,
			xmax=110,
			xlabel={Mean of two measures},
			every outer y axis line/.append style={black},
			every y tick label/.append style={font=\color{black}},
			ymin=-30,
			ymax=50,
			ylabel={Difference between two measures},
			ylabel near ticks, 
			]
		\addplot [color=darkgray,mark size=1pt,only marks,mark=*,mark options={solid}] table[x=mean,y=diff]{result/bland_motion_data_old_read.txt};
		
		\addplot [color=purple,mark size=1pt,only marks,mark=square*,mark options={solid}] table[x=mean,y=diff]{result/bland_motion_data_old_watch.txt};
		
		\addplot [color=teal,mark size=1pt,only marks,mark=triangle*,mark options={solid}] table[x=mean,y=diff]{result/bland_motion_data_old_talk.txt};
		
		\addplot [color=black,solid,forget plot] coordinates {(40,\DoldHRMDmotion) (110,\DoldHRMDmotion)};
		\addplot [color=black,dashed,forget plot] coordinates {(40,\DoldHRMDminusSDmotion) (110,\DoldHRMDminusSDmotion)};
			  
		\node at (axis cs: 95,-25.5) {mean $- 1.96$SD };
		\addplot [color=black,dashed,forget plot]  coordinates {(40,\DoldHRMDplusSDmotion) (110,\DoldHRMDplusSDmotion)};
				 
		\node at (axis cs: 95,+40) {mean $+ 1.96$SD };
				
		\end{axis}
		\end{tikzpicture}
		\caption{PR estimate performance using \oldMethod}
		
	\end{subfigure}
	\caption{Bland-Altman plot: comparison of PR derived from camera PPG and from ground truth \pulseox\ for $5$ subjects having different skin tones under three motion scenarios --- (i)Reading text, (ii) watching video, (iii) talking. }
	\label{fig:motion_bland}

\end{figure}

The improvement in SNR under various motion scenarios reduces the error in PR estimation. Figure~\ref{fig:motion_bland} shows the agreement between camera derived PR and ground truth \pulseox\ derived PR from $5$ subjects under the three motion scenario (reading, watching, talking) using Bland-Altman plot. For PPG estimated using \oldMethod, the mean bias $\bar{d} = \DoldHRMDmotion$ bpm with $95\%$ limit of agreement between $\DoldHRMDminusSDmotion$ to $\DoldHRMDplusSDmotion$ bpm. Using the \mrc\ reduces the error in PR estimate to $\bar{d} = \DmrcHRMDmotion$ bpm with $95\%$ limit of agreement between $\DmrcHRMDminusSDmotion$ to $\DmrcHRMDplusSDmotion$ bpm. Note, \algo\ do not perform well for PR estimation under talking scenario possibly because of large non-rigid motion. For non-talking motion scenario (reading+watching), the  error in PR estimate is $\bar{d}=\DmrcHRMDreadwatch$~bpm with  $95\%$ limit of agreement being $\DmrcHRMDminusSDreadwatch$ to $\DmrcHRMDplusSDreadwatch$~bpm which is much better. Thus, one can possibly use computer vision based activity recognition to not estimate PR during large non-rigid motion.

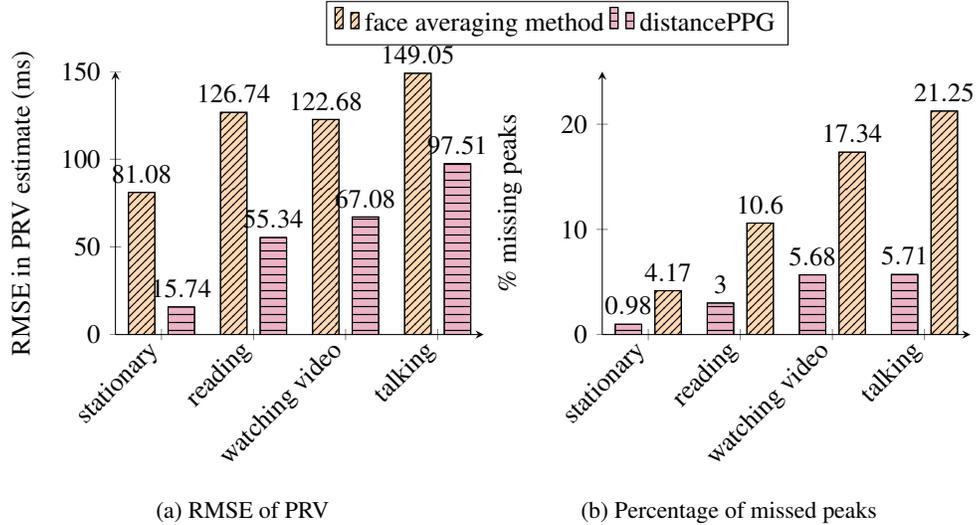
\begin{figure}[h]
\centering

	\begin{subfigure}[t]{0.48\textwidth}
	\centering 
		\begin{tikzpicture}[baseline]

		    \begin{axis}[
		    	ybar=5pt,
		    	width=2.55in,
		    	height =2in,
		    	xtick={1,2,3,4},
		    	axis x line = bottom,
		    	axis y line = left, 
		    	ymin=0,		    			   
		    	ymax=150,
		    	xmin=0.5,
		    	xmax=4.5,
		    	nodes near coords,
			   	nodes near coords align={vertical},
		    	xticklabels={stationary, reading, watching video,talking},
		    	x tick label style={rotate=45,anchor=east},
		        ylabel=RMSE in PRV estimate (ms),
		       legend style={at={(1.2,1.1)},anchor=south},       			      legend columns=2,]
		   \addplot[black, fill=orange!30!white,
		    	mark=none,
		    	postaction={pattern=north east lines},] table[x=IDs,y=OLD] {result/IBI_RMSE_motion.txt};
		    \addplot[black,
		    		    		  fill=purple!30!white,
		    		    		  mark=none,
		    		    		  postaction={pattern=horizontal lines},
		    		    		  ] table[x=IDs,y=MRV] {result/IBI_RMSE_motion.txt};
		    \legend{\oldMethod \qquad,\algo}
		    \end{axis}
		\end{tikzpicture}
		\caption{RMSE of PRV}
	\end{subfigure} \begin{subfigure}[t]{0.48\textwidth}
	\centering												\begin{tikzpicture}[baseline]

		    \begin{axis}[
				ybar=5pt,
		    	width=2.55in,
		    	height =2in,
		    	xtick={1,2,3,4},
		    	axis x line = bottom,
		    	axis y line = left, 
		    	ymin=0,		    			   
		    	ymax=25,
		    	xmin=0.5,
		    	xmax=4.5,
		    	nodes near coords,
			   	nodes near coords align={vertical},
		    	ylabel= \% missing peaks, 
		    	xticklabels={stationary, reading, watching video,talking}, 
		    	x tick label style={rotate=45,anchor=east},]
		   \addplot[black, fill=purple!30!white,
		   mark=none,
		   postaction={pattern=horizontal lines},
		   ] table[x=IDs,y=MRV] {result/missing_motion.txt};
			\addplot[black, fill=orange!30!white,
		    	mark=none,
		    	postaction={pattern=north east lines},] table[x=IDs,y=OLD] {result/missing_motion.txt};
		     \end{axis}
		\end{tikzpicture}
		\caption{Percentage of missed peaks}
	\end{subfigure} 
	\caption{Pulse rate variability (PRV) estimation performance under three motion scenario: (i) running, (ii) watching, (iii) talking  for $5$ subjects of varying skin tones.  (Left) Bar-plot  shows the RMSE in the timings of the peaks in the camera-based PPG waveform when compared with \pulseox\ based ground truth. (Right) bar-plot shows the percentage of missing peaks in camera-based PPG waveform.}
	\label{fig:motion_IBI}
\end{figure}

Figure~\ref{fig:motion_IBI} shows the RMSE in PRV estimation and the corresponding percentage of missing peaks in camera-based PPG for the three motion scenario along with  stationary base case. \Algo\ provides  reduction in RMSE of PRV (by around 1.5x) and the percentage of missing peaks (by atleast by 2x) when compared to \oldMethod\ under motion scenarios. Still, there is scope for further improvement as the RMSE in PRV estimate is around $50-100$~ms, and one can achieve around $16.6$~ms RMSE with the $30$~fps camera. These errors are due to large non-rigid motion of the face during smiling, talking etc.

\subsection{Performance under various lighting conditions}
Figure~\ref{fig:SNR_light} shows the SNR of camera-based PPG estimate under different lighting condition ($50-650$~lux) using both \algo\ (marked as squares) and \oldMethod\ (marked as circles) for two subjects having pale white (dashed line) and brown skin (solid line) tones. On an average \algo\ provides an SNR gain of $\DSNRGainLightW$~dB for pale white person, and an SNR gain of $\DSNRGainLightB$~dB for a brown skin tone person. Though we tried to keep all other experimental parameter (particularly motion) invariant (static in this case) while varying the light intensity, one should not read much from the ups and downs in the plot, as they mostly are due to uncontrolled experimental variations. Nonetheless, the general trend is that SNR of camera-based PPG estimate increases as we increase the light intensity. This trend is more evident for  \algo\ compared to \oldMethod which shows that \algo~is more immune to such experimental variations. 

\begin{figure}[h]
\centering
	\begin{tikzpicture}
		\begin{axis}[%
					name=lighting,
					width=3.0in,
					height=1.5in,
					separate axis lines,
					scale only axis,
					xlabel={Ambient light intensity (lux)},
					xmajorgrids,
					ylabel={SNR (dB)},
					ymajorgrids,
					legend style={at={(0.5,1.1)},anchor=south,legend cell align=left,draw=black}, 
   				   legend columns=2, 
					]
		\addplot[teal, dashed, mark = *, mark size=2pt] table [x=light, y=SNR_MRC_U1]{result/SNR_light.txt};
		\addlegendentry{fair skin tone - \algo}
		\addplot[purple, dashed, mark = square*, mark size=2pt] table [x=light, y=SNR_OLD_U1]{result/SNR_light.txt};
		\addlegendentry{fair skin tone - \oldMethod }
		\addplot[teal, solid, mark = *, mark size=2pt] table [x=light, y=SNR_MRC_U2]{result/SNR_light.txt};
		\addlegendentry{dark skin tone - \algo }
		\addplot[purple, solid, mark = square*, mark size=2pt] table [x=light, y=SNR_OLD_U2]{result/SNR_light.txt};
		\addlegendentry{dark skin tone - \oldMethod }
		\end{axis}

		\end{tikzpicture}
	\caption{SNR of camera-based PPG using \algo\ and \oldMethod\ under various lighting conditions. \Algo\ provides on an average $\DSNRGainLightB$~dB SNR improvement for brown skin tone (squares), and provides on an average $\DSNRGainLightW$~dB SNR improvement for fair skin tone person under various lighting conditions}
	\label{fig:SNR_light}
\end{figure}

\subsection{Comparison with ICA based method}
For static scenario ($4$ subjects having non-Caucasian skin tones), the mean SNR of camera-based PPG using \algo\ is $\DSNRmrcStatic$~dB, and using ICA-based method is $\DSNRICAStatic$~dB. The mean error in PR estimation is $\DMeanHRErrormrcStatic$~bpm and RMSE of PRV estimation is $\DRMSEIBImrcStatic$~ms for \algo, and corresponding errors for ICA-based method is $\DMeanHRErrorICAStatic$~bpm and $\DRMSEIBIICAStatic$~ms respectively. Thus, the small improvement in SNR ($1$~dB) by using \algo\ results in better performance in PRV estimation accuracy.  

For talking scenario ($3$ subjects under $500$~lux illumination, $1$ subject under $300$~lux ), both \algo\ and ICA-based method performs poorly,  possibly because of large motion artifact which renders small PPG signal estimation difficult. The mean SNR of camera-based PPG are $\DSNRmrcMotion$~dB and $\DSNRICAMotion$~dB for \algo\ and ICA-based method respectively. Due to such low SNR, the mean error in pulse rate  suffers for both methods, and is $\DMeanHRErrormrcMotion$~bpm and $\DMeanHRErrorICAMotion$~bpm for \algo\ and ICA-based method respectively.

\section{Discussion}
\subsection{Goodness metric as signal quality index}
We used goodness metric as a substitute for the signal quality or SNR of the PPG  signal obtained from different regions of the face. To understand how good our goodness metric substitute is, we compute scatter plot between goodness metric $G_i$ in dB and SNR of the PPG  signal $\hat{y}_i(t)$  obtained from different ROI $\mathcal{R}_i$ inside the face. Figure~\ref{fig:goodnessSNR} shows this scatter plot for two subjects having white/fair and brown/dark skin tones.

\definecolor{mycolor1}{rgb}{0.84706,0.16078,0.00000}%

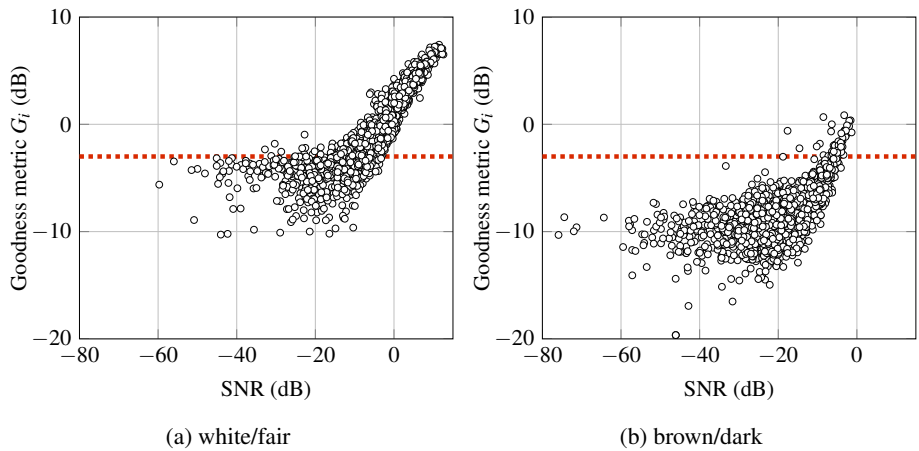
\begin{figure}
\centering
	\begin{subfigure}[b]{0.45\textwidth}
		\begin{tikzpicture}[scale=0.85,baseline]
		\begin{axis}[%
			scale only axis,
			width=2.3	in,
			xmin = -80,
			xmax = 15,
			ymin = -20,
			ymax = 10,
			every outer x axis line/.append style={black},
			every x tick label/.append style={font=\color{black}},
			xlabel={SNR (dB)},
			xmajorgrids,
			every outer y axis line/.append style={black},
			every y tick label/.append style={font=\color{black}},
			ylabel={Goodness metric $G_i$ (dB)},
			y label style={at={(axis description cs:0.1,.5)},anchor=south},
			ymajorgrids,]
			\addplot[only marks,mark=*,mark options={},mark size=1.5000pt,draw=black,fill=white]
			table[x=SNR,y=goodness]{result/scatterSNRGoodnessUID9.txt};
			\addplot [color=mycolor1,dotted,line width=2.0pt,forget plot]
			table[row sep=crcr]{%
					-80	-3\\
					15	-3\\
					};
		\end{axis}

		\end{tikzpicture}
 		\caption{white/fair}
  		\label{fig:fair}
	\end{subfigure}~ \begin{subfigure}[b]{0.45\textwidth}
		\begin{tikzpicture}[scale=0.85,baseline]
			\begin{axis}[%
			scale only axis,
			width=2.3in,
			xmin = -80,
			xmax = 15,
			ymin = -20,
			ymax = 10,
			every outer x axis line/.append style={black},
			every x tick label/.append style={font=\color{black}},
			xlabel={SNR (dB)},
			xmajorgrids,
			every outer y axis line/.append style={black},
			every y tick label/.append style={font=\color{black}},
			ylabel={Goodness metric $G_i$ (dB)},
			y label style={at={(axis description cs:0.1,.5)},anchor=south},
			ymajorgrids,]
			\addplot[only marks,mark=*,mark options={},mark size=1.5000pt,draw=black,fill=white]
			table[x=SNR,y=goodness]{result/scatterSNRGoodnessUID12.txt};
			\addplot [color=mycolor1,dotted,line width=2.0pt,forget plot] table[row sep=crcr]{%
								-80	-3\\
								15	-3\\
								};
			\end{axis}
			\end{tikzpicture}
	 		\caption{brown/dark}
	  		\label{fig:dark}
	\end{subfigure}
	\caption{Scatter plot between goodness metric (dB) and SNR (dB): Goodness metric defined in this paper is a good substitute for signal quality (SNR) in regions having goodness metric greater than $-3$~dB (linear region of the scatter plot). }
	\label{fig:goodnessSNR}
\end{figure}

The scatter plot shows that goodness metric is a good substitute for SNR in regions having goodness metric greater than $-3$~dB. For regions having lower goodness metric, goodness metric overestimates the SNR. Thus, we also reject regions having very low goodness metric score ($G_i<-6$~dB or $G_i<0.25$) in the PPG estimation Equation~\eqref{eg:PPG_estimate}. Completely rejecting regions having very small goodness metric value $G_i<0.25$ slightly improve the overall SNR of camera-based PPG signal (by $0.2$~dB). 

Our definition of goodness metric exploited a known prior structure in photoplethysmogram signals, i.e. PPG signals are periodic with dominant frequency equal to the pulse rate, and developed a signal quality index based on the spectrum of the PPG signal extracted from a region. This signal quality index helped us propose weighted averaging algorithm to improve the SNR of the overall PPG estimate. Thus, we broke down the problem of robust camera-based PPG waveform estimation into two stages: (i) Estimation of dominant frequency (PR), which is a relatively easy problem, and (ii) robust estimation of PPG waveform by defining a new signal quality index based on the pulse rate.

It is important to note that merely filtering the PPG signal in a narrow frequency band around the pulse rate ($[\text{PR}-b,\text{PR}+b]$~Hz), a method generally used for weak periodic signal detection, would not give a good estimate of the PPG waveform, as the spectral band of interest for PPG signal is much wider (\filterBand). In other words, we care about the exact shape of the PPG waveform (e.g. inter-beat-interval information, systolic and diastolic peaks etc), and thus we should design an estimation algorithm which preserves the shape. One may find similar opportunities and concerns in many other periodic biological signal estimation problems e.g. electrocardiogram (EKG), event related potentials (ERP), etc. Goodness metric based signal quality index developed in this paper can be expanded for these scenarios as well, particularly when one records these biological signals using sensors distributed in space and wants to develop a technique to weigh the sensor based on their signal quality.

\subsection{Goodness metric for blood perfusion}

The goodness metric profile over a face depends on two factors - (i) spatial illumination profile over the face $I(x,y)$. and (ii) strength of modulation ($\alpha_i$) of reflected light which depends on the blood perfusion profile. As the illumination profile can change over time, \algo\ re-estimates the goodness metric every $T$ seconds. Figure  \ref{fig:MRC_Algorithm} of goodness metric overlay clearly shows that forehead region is best for extracting PPG signal from the face. Our finding is in agreement with the findings of other researchers that forehead represent a suitable site for camera-based PPG estimation. This is because the forehead region have much better blood perfusion compared to other regions \cite{fernandez_evaluation_2007}. 

Further, it might be feasible to develop a new metric for the blood perfusion profile over the face if one can separately estimate the spatial illumination profile and compensate for it. Blood flow in peripheral tissue is generally estimated using either laser Doppler flowmetry or using laser speckle contrast imaging (LSCI) techniques. Camera-based blood flow monitoring, if developed, will be highly useful. 

\subsection{PRV estimation using camera PPG}
The RMSE of PRV that we could achieve for fairer skin tones under stationary scenario is around $16-20$ ms. It is difficult to achieve something significantly below this error value using a $30$~fps camera-based PPG signal. This is because the time difference between two camera frames is $33.3$~ms, and thus there is an ambiguity of $\pm16.6$~ms in peak detection. No amount of interpolation (we interpolated camera PPG to $500$~Hz using spline interpolation) can completely recover the higher frequency details lost because of the low sampling rate. 

One solution to further reduce the RMSE of PRV is to use higher frame rate camera (e.g., $100$~fps). Since maximum achievable exposure time of higher frame rate cameras would be smaller (e.g. $10$~ms), this can reduce  the amount of light entering the camera sensor, and thus the SNR of the estimated camera PPG signal as well. This will make faithful pulse minima detection, an essential step for PRV estimation, difficult. The algorithmic SNR gain achieved due to \algo\ can be useful here,  as  one can compensate the SNR loss because of higher frame rate with the algorithmic SNR gain due to \algo.

Pulse rate variability (PRV) is one of the most commonly used measure in psychophysiology research, and is employed broadly as a determinant of the status of the autonomic nervous system (ANS) \cite{heathers_smartphone-enabled_2013}. Camera-based PRV estimation can enable researchers to conduct large scale psychophysiological studies of statistical significance outside their lab settings. One needs to evaluate and understand the impact of error in camera-based PRV on various metric of interest in psychophysiology such as high frequency (HF) power, low frequency (LF) power, LF/HF power ratio of PRV, standard deviation of all normal heart period (SDNN) etc. This will be one of the future direction of our research.

\section{Conclusion}
In this paper, we dived deeper into understanding the challenges involved in estimating photoplethysmogram waveform using a camera-based system. We developed a new method, \algo, that addresses these challenges. We evaluated our algorithm on people having diverse skin tones (pale/white to brown/dark), under various lighting conditions ($50$~lux to $650$~lux) and natural motion scenarios, and showed significant improvement in the accuracy of vital sign (pulse rate, pulse rate variability) estimation. Our major contribution is to develop a formal method to take into consideration differences in blood perfusion and incident light intensity in different regions of the face to improve the accuracy of vital sign estimation under difficult scenario (e.g. dark skin tones and/or low lighting condition). Further, we proposed a new motion-tracking algorithm which tracks different regions of the face separately, to improve performance of vital sign estimation under natural motion scenarios like reading content on computer screen, watching video, and talking over skype. 

We also highlighted limitations of \algo\ algorithm, e.g. during talking scenario, where vital sign estimation accuracy suffers. The major challenge in such scenario is significant variations in skin surface reflectance component during large motion which falls in the frequency band of interest of PPG signal (\filterBand). The small changes in skin subsurface reflectance, which encodes the PPG signal, is not recoverable in presence of large in-band surface reflectance changes. Thus, the general approach we adopted in \algo\ is to reject regions undergoing large surface reflectance changes. Consequently, during large motion we end up rejecting a majority of face region and thus our estimate of PPG signal was inaccurate. 

Two popular apps for measuring non-contact pulse rate using the color change (or PPG) signal from a person's face are Philips Vital sign camera \cite{_philips_????} and Cardiio \cite{_cardiio_????}. Both these apps require users to be at rest facing the camera in a well lit environment to be effective \cite{_cardiio_????}. The \algo\ algorithm discussed in this paper address these challenges and thus would extend the use case of mobile phone and computer apps for vital sign monitoring. We are in the process of developing a realtime PC-based application to robustly estimate PPG signal using a webcam. Our future work includes porting our code to popular mobile platforms (Android/iOS), and further improving the performance of \algo\ under motion scenarios. 

\section*{Acknowledgment} 
This work was partially supported by NSF CNS 1126478, NSF IIS-1116718, Rice University Graduate Fellowship, Texas Instruments Fellowship, and Texas Higher Education Coordinating Board: THECB-NHARP 13308
\end{document}